%% file: main.tex
\documentclass[11pt]{article}

\usepackage[utf8]{inputenc}
\usepackage[T1]{fontenc}

\usepackage{arxiv}

\usepackage{amsmath}
\usepackage{amssymb}
\usepackage{graphicx}
\usepackage{booktabs}
\usepackage{subcaption}
\usepackage{enumitem}
\usepackage{tcolorbox}

\usepackage{hyperref}
\usepackage{url}
\hypersetup{
  colorlinks=true,
  linkcolor=arxivaccent,
  citecolor=arxivaccent,
  urlcolor=arxivaccent,
  filecolor=arxivaccent,
  breaklinks=true,
}

\newtcolorbox{finding}{
  colback=arxivaccent!7,
  colframe=arxivaccent,
  boxrule=0.5pt, arc=2pt,
  left=8pt,right=8pt,top=4pt,bottom=4pt,
  before skip=8pt,after skip=8pt,
}

\title{The Latent Bridge: A Continuous Slow--Fast Channel for Real-Time Game Agents}

\author{%
  \begin{tabular}{@{}c@{\hspace{4em}}c@{}}
    Bojie Li & Noah Shi \\
    Pine AI & University of Washington
  \end{tabular}%
}
\date{}
\runningtitle{The Latent Bridge: A Continuous Slow--Fast Channel for Real-Time Game Agents}

\begin{document}

\maketitle

\begin{abstract}
A real-time agent for general computer use---with games as the most demanding case---must act within tens of milliseconds while still planning over seconds. These two regimes sit at opposite ends of the latency--quality tradeoff. A \emph{reasoning} VLM (Qwen3-VL-8B-Thinking) deliberates effectively but requires $\sim$1.5\,s per response---far too slow for a 15\,Hz control loop. In contrast, a \emph{reactive} VLM (MiniCPM-o~4.5) acts in milliseconds but underperforms on planning-heavy tasks. We couple two frozen models of matched scale (9\,B reactive, 8\,B reasoning), leaving the communication \emph{channel} as the sole trainable component. The standard coupling is a \textbf{Text Bridge} ($T$): the slow model writes a suffix the fast model reads. We introduce a learned continuous \textbf{Latent Bridge} ($L$) that projects the slow model's residuals into the fast model's input-embedding space in a LLaVA-style manner, avoiding any text round-trip; both are compared against \textbf{Fast-Only} ($F$). On 7 Atari games and a driving domain (MetaDrive), tuning the action decoder per channel on held-out seeds, the Latent Bridge matches or beats the Text Bridge in every domain: it significantly improves two games (MsPacman $+57\%$, RoadRunner $+28\%$) and is a safe drop-in elsewhere. Combining both channels interferes destructively (RoadRunner $-96\%$), so only one should be used. The benefit is highly predictable: the bridge helps if and only if slow reasoning already beats fast reaction ($T\!>\!F$)---the Latent and Text gains over Fast-Only move together at $r=0.93$. MetaDrive is the controlled negative, where the Latent Bridge is demonstrably inert because the Text Bridge adds no value. We release replay recordings and reproducible pipelines.
\end{abstract}

\begin{center}
\small
Code: \url{https://github.com/19PINE-AI/latent-bridge-games} \\[2pt]
Website: \url{https://01.me/research/latent-bridge-games}
\end{center}
\vspace{-0.6em}

\section{Introduction}

We seek agents for \emph{general computer use} (GCU): read the screen, issue low-level inputs, close the loop, repeat. Its most demanding form is real-time video games on a phone or desktop, where the agent reacts to raw pixels every few tens of milliseconds while pursuing goals that need planning over seconds---two opposing regimes, fast reaction and slow deliberation, that are hard to get from one model without purpose-built training.

\begin{figure}[t]
\centering
\includegraphics[width=\textwidth]{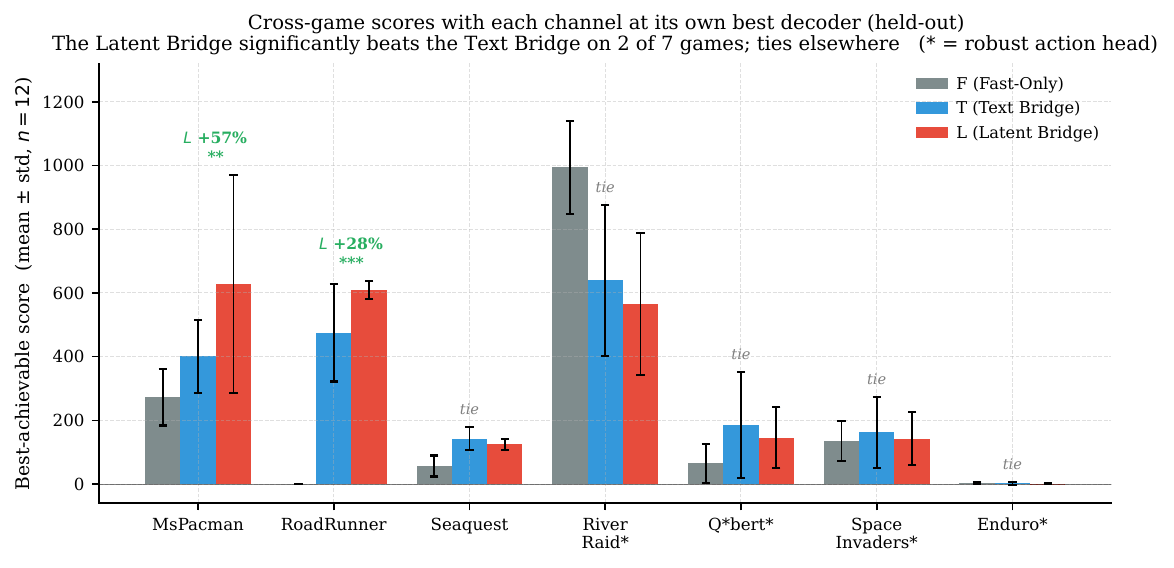}
\caption{\textbf{Cross-game scores with each channel at its own best decoder, selected on held-out seeds} ($n=12$; reported action-head variant per game, selected by the rule in \S\ref{sec:setup_variant}). The Latent Bridge significantly beats the Text Bridge on MsPacman ($+57\%$) and RoadRunner ($+28\%$) and ties on the other five, never losing. Stars: $L$-vs-$T$ significance by Mann--Whitney $U$ (\texttt{**}\,$p<.01$, \texttt{***}\,$p<.001$); $*$ on a game label marks the robust-action-head variant (\S\ref{sec:ood}). Both variants are tabulated in Appendix~\ref{app:full}.}
\label{fig:headline}
\end{figure}

\paragraph{Why couple two models rather than train one.} A single model \emph{can}, in principle, do both---StepFun's open-weights \emph{Step-Audio-R1.1}~\cite{stepaudior1} interleaves reasoning with speech, answering fast while it keeps thinking---but training one to do both well chases a moving target. Reasoning models~\cite{openai_o1,deepseek_r1} advance every few months: GPT-5.2 through 5.5~\cite{gpt52} and Claude Opus 4.5 through 4.8~\cite{claudeopus45,claudeopus48} all shipped within about six months, in a race won on coding~\cite{swebench} and agentic tool use~\cite{taubench}, not real-time interaction. Building interaction in as its own objective---and creating the data it demands---risks diluting the raw intelligence the race optimizes for, the capability that matters most. Even frontier labs can ill afford that; and at the smaller sizes others can train, no tuning closes the gap to the state of the art.

\paragraph{The decoupled alternative.} The pragmatic alternative is to leave the best reasoning model untouched and pair it with a separate, custom-trained model for the real-time loop. Recent decoupled systems do exactly this---Thinking Machines Lab's \emph{Interaction Models}~\cite{interaction_models} couple a live interaction model to an asynchronous background reasoner, as do xAI's \emph{Grok Voice Think Fast}~\cite{grokvoice} and Pine~AI's voice agent~\cite{pineai}---running a fast responder in the loop and a slow reasoner in the background, coupled by streamed text or context. These are voice-centric, tightly co-designed, and largely closed.

\paragraph{Our setting.} We carry this principle to general computer use, and specifically to real-time games. The open, general-purpose models one would actually deploy come with no built-in fast/slow split: a \emph{reasoning} VLM (Qwen3-VL-8B-Thinking~\cite{qwen3vl}) deliberates well but needs $\sim$1.5\,s per response---tens of frames late for a $\sim$15\,Hz loop---while a \emph{reactive} VLM (MiniCPM-o~4.5~\cite{minicpmo}) answers in milliseconds but plans poorly: acting alone (\textbf{Fast-Only}, $F$) it leaves much on the table---on MsPacman, coupling in a reasoner roughly doubles the score. We therefore couple two \emph{frozen, open-source} models of matched scale and study the question these systems leave implicit: \emph{how} should the slow model's deliberation reach the fast model---can a learned continuous latent channel beat the standard text coupling?

\begin{figure}[t]
\centering
\includegraphics[width=\textwidth]{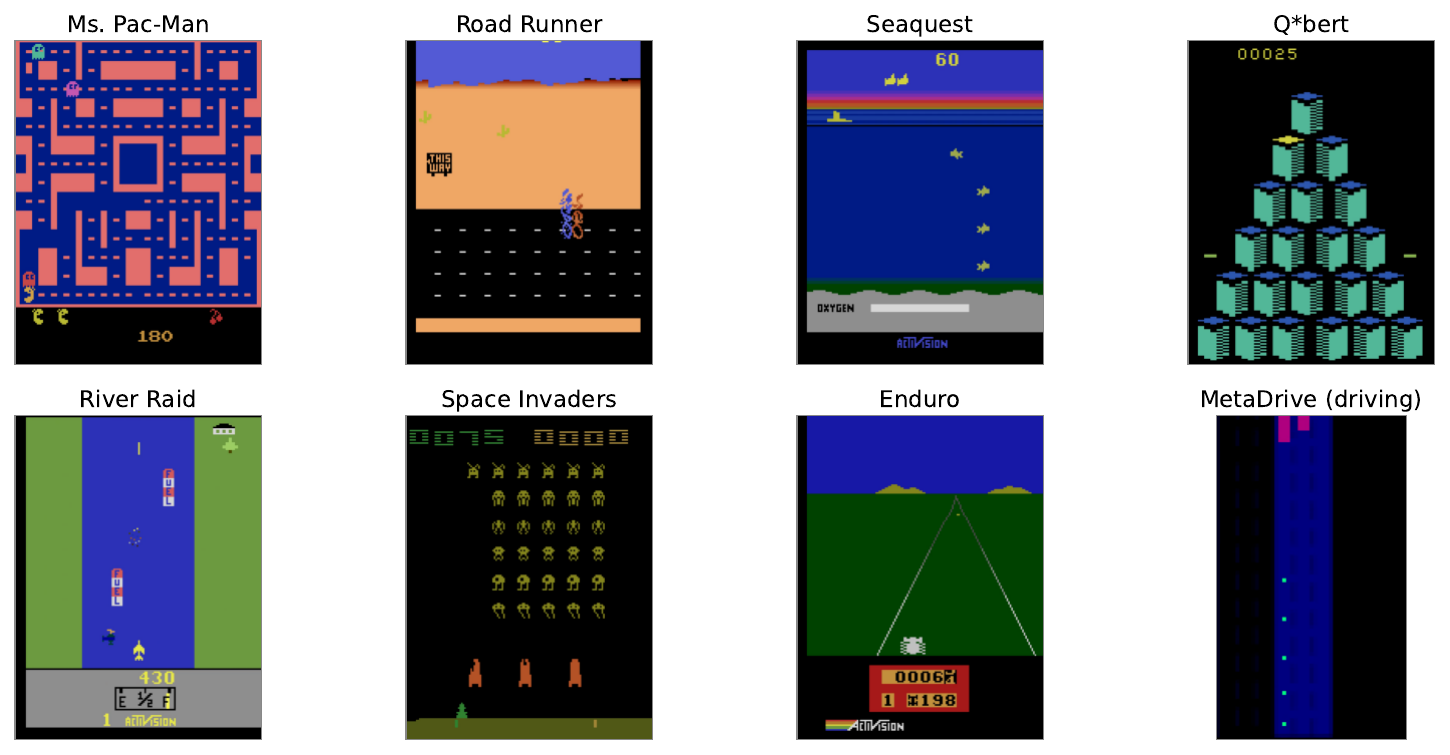}
\caption{\textbf{The eight evaluated domains.} Seven Atari games (raw pixels, $\sim$15\,Hz
control), spanning fast hazard-avoidance---ghosts, obstacles, enemy fire---and slower
route/strategy planning, plus MetaDrive (top-down driving), our non-Atari controlled negative.
Frostbite (excluded) and Pong (reported but uninformative) are not shown.}
\label{fig:games}
\end{figure}

\paragraph{Text Bridge vs.\ Latent Bridge.} The standard coupling is the \textbf{Text Bridge} ($T$): the slow model writes its conclusion and the fast model reads it as a prompt suffix. We introduce the learned continuous \textbf{Latent Bridge} ($L$): it projects the slow model's residual stream directly into the fast model's input-embedding space (LLaVA-style) and prepends a few latent tokens, with no text round-trip. Both are compared against \textbf{Fast-Only} ($F$). Because both base models are frozen and matched in scale (9\,B fast, 8\,B slow), the \emph{channel}---the sole learned component---is the only variable under study. The fast model acts at $\sim$15\,Hz, the slow model deliberates at $\sim$1\,Hz. We evaluate on Atari~\cite{atari} (100\,ms ghost-dodging meets 10\,s route-planning) and the MetaDrive driving simulator as a controlled negative. Figure~\ref{fig:games} previews all eight evaluated domains.

\paragraph{What this paper shows.}
\begin{itemize}[topsep=0pt,itemsep=2pt,leftmargin=*]
\item \textbf{Architecture matters.} A first attempt (v1, 256-d cross-attention at 2 of 36 layers) converged to KL=0.004 offline yet \emph{failed} at deployment; the working v2 is the LLaVA pattern---project slow residuals into the fast LLM's 4096-d input-embedding space and prepend, so all layers attend via standard causal attention (\S\ref{sec:arch}).
\item \textbf{Tuned per channel, the Latent Bridge is never significantly worse than the Text Bridge and is significantly better on 2 of 7 games} (MsPacman $+57\%$, RoadRunner $+28\%$; Figure~\ref{fig:headline}, \S\ref{sec:best_achievable}). A fixed greedy decoder (always taking the highest-probability action) flatters the Latent Bridge to 4 wins, but that edge is \emph{greedy-specific}---it vanishes once actions are instead sampled (\S\ref{sec:qbert}). Since the decoder is a tunable deployment hyperparameter, the fair test gives each channel its own best decoder, selected on held-out seeds; the two in fact prefer \emph{different} decoders.
\item \textbf{Using both channels at once \emph{hurts}---couple via exactly one channel} (\S\ref{sec:combined}). Feeding text suffix and latent tokens in one pass never beats the better single channel and significantly interferes on 3 games (RoadRunner $-96\%$): the frozen head, trained on one conditioning signal at a time, gets a worse policy from two. The Latent Bridge is the safe single-channel default.
\item \textbf{Whether the bridge helps is a property of the task, not the channel.} Across 7 Atari games and a driving domain (MetaDrive), the Latent Bridge's benefit over Fast-Only ($L\!-\!F$) tracks the Text Bridge's ($T\!-\!F$) at Pearson $r=0.93$: the bridge pays off exactly when slow reasoning beats reaction ($T\!>\!F$), and is inert or harmful otherwise. MetaDrive is the controlled negative: swapping the trained latent for zeros or random vectors (a \emph{bridge-replacement control}) leaves the score unchanged, confirming the Latent Bridge is inert there (\S\ref{sec:predictor}). We frame it this way rather than as a ``text is bandwidth-limited'' story, which our own ablations do not support.
\item \textbf{Most collapsed cells are not bridge failures} but out-of-distribution (OOD) brittleness of the fast model's action head under the suffix/bridge inputs. Retraining that head to tolerate them fixes the collapses, dramatically so on River~Raid (\S\ref{sec:ood}).
\item \textbf{The Latent Bridge does not always beat Fast-Only.} On several games where it beats the Text Bridge, Fast-Only still beats both: the bridge improves over text \emph{at matched architecture}; it does not always justify running a slow model at all.
\end{itemize}

\section{Setup}
\label{sec:setup}

\paragraph{Models.} We use two frozen multimodal models of similar scale. The \emph{fast} (reactive) model is MiniCPM-o 4.5~\cite{minicpmo} (9\,B parameters, bf16). The \emph{slow} (reasoning) model is Qwen3-VL-8B-Thinking~\cite{qwen3vl} (8\,B parameters, bf16).

\paragraph{Tick budgets.} Atari environments run at 60\,Hz but the fast model is limited to a 15\,Hz control rate---one action every $\sim$67\,ms (a \emph{tick}). The slow model produces one deliberation (an \emph{emission}) asynchronously at roughly 1\,Hz, so each emission is reused for $\sim$15 ticks; because its residuals are cached, the projected latent tokens are identical across those ticks until the next emission lands. Measured warm-path inference latencies (with vision caching) are $F=33$\,ms for the fast model alone and $L\approx 38$\,ms when using the Latent Bridge (8 prepended tokens add $\sim$5\,ms). End-to-end wall-clock time for the full Latent-Bridge system is dominated by GPU contention from the asynchronous slow model rather than by the bridge itself. (Our headline scores in \S\ref{sec:results} use per-tick vision rather than this cached path, so they are correctness numbers, not latency-optimized; see Appendix~\ref{app:latency}.)

\begin{figure}[t]
\centering
\includegraphics[width=\textwidth]{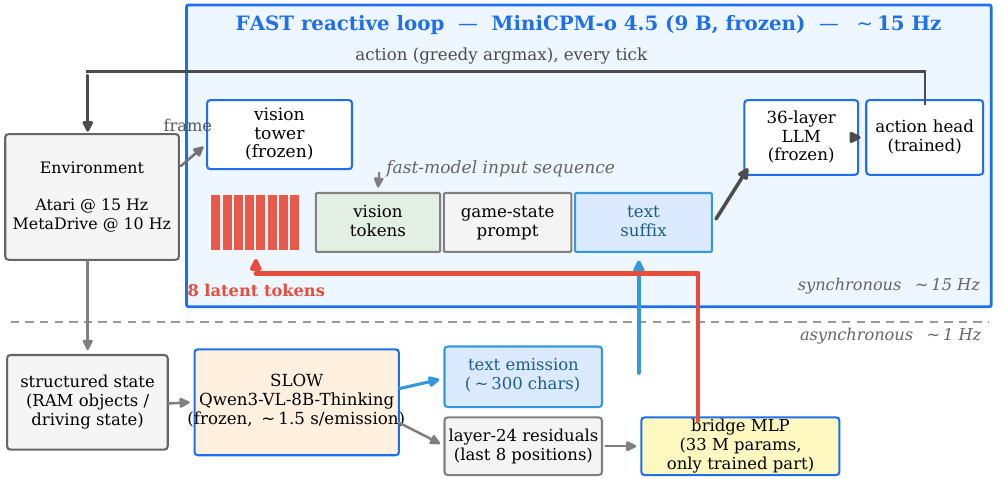}
\caption{\textbf{System architecture.} The fast model (MiniCPM-o 4.5, frozen) runs the reactive loop: vision tokens and a game-state prompt feed a 36-layer LLM whose trained action head emits one action per tick ($\sim$33--38\,ms warm path). The slow model (Qwen3-VL-8B-Thinking, frozen) reasons asynchronously over structured state at $\sim$1\,Hz; the fast loop never blocks on it and reuses the latest emission until replaced. The slow output reaches the fast model two ways: its text emission appended as a prompt suffix, and its layer-24 residuals projected through the 33\,M-parameter bridge MLP---the only trained component---into 8 latent tokens prepended to the input. The three strategies (Fast-Only, Text Bridge, Latent Bridge) toggle which of these the fast model receives.}
\label{fig:system}
\end{figure}

\paragraph{Three strategies compared (Figure~\ref{fig:system}).}
All three share the frozen fast model and its action head; they differ only in
what the slow model contributes to the fast context.
\begin{itemize}[topsep=2pt,itemsep=2pt,leftmargin=*]
\item \textbf{Fast-Only} ($F$): the fast model acts alone, ignoring the slow model.
\item \textbf{Text Bridge} ($T$): Fast-Only plus the slow model's full text emission
appended verbatim as a prompt suffix (median 302 chars; samples in
Appendix~\ref{app:emissions}).
\item \textbf{Latent Bridge} ($L$): Fast-Only plus the slow model's projected latent
tokens prepended (8 tokens, 4096-d each).
\end{itemize}
We write the three in prose by name and abbreviate them $F$, $T$, $L$ in figures and
formulas. (A slow-\emph{only} policy is not a fourth strategy: it must finish a full
perception--reasoning--emission cycle synchronously before each action, so at the
slow model's $\sim$1.5\,s per emission, it acts at well under 1\,Hz, missing tens of
frames per decision---the speed problem that motivates the coupling.)

\paragraph{Training pipeline.}
Three stages, in order (only the bridge MLP is newly trained; both VLMs stay frozen):
\begin{itemize}[topsep=3pt,itemsep=2pt,leftmargin=*]
\item \textbf{Action-head cloning}: behavioral cloning (BC) from Stable-Baselines3 (SB3) experts~\cite{sb3zoo} onto the fast model's \emph{action head}---the small head mapping the LLM's final hidden state to a game action.
\item \textbf{Trajectory caching}: collect Text-Bridge trajectories and cache, per tick, (frame, slow text, slow residuals at layer 24, last 8 positions).
\item \textbf{Bridge distillation}: train the projection MLP to minimize $\mathrm{KL}(\pi_L\,\|\,\pi_T)$---making the latent-conditioned policy match the text-conditioned one---with both base models frozen ($\sim$5K samples/game, final KL $\sim$0.005). Hyperparameters and per-game val-accuracies are in Appendix~\ref{app:training}.
\end{itemize}

\paragraph{Action-head robustness is a tuned hyperparameter (\emph{not} per-game cherry-picking).}
\label{sec:setup_variant}
For each game, we train two variants of the action head: \emph{bare} (trained on plain 
state prompts) or \emph{robust} (trained with suffix-probability 0.5: half of the batches get a 
slow-style text suffix appended to teach the head to tolerate the OOD suffix/bridge inputs it 
meets at deployment; \S\ref{sec:ood}). This is a single binary hyperparameter. We select the 
variant per game by the following rule: choose the variant with the higher $L$ performance, 
provided its Text Bridge has not collapsed ($T\!>\!0$, so the $L$-vs-$T$ comparison is
meaningful and not a tie at zero). Ties and double-collapses are broken in favor of higher $T$ performance.
This rule is \emph{conservative}. On Enduro, the bare variant has marginally higher $L$ ($7.8$ 
vs $5.8$) but $T\!=\!0$, so we select the robust variant ($T\!=\!5$, $L\!=\!6$). This choice
lowers, rather than raises, the reported $L$ for that game. Because the rule maximizes $L$---
the deployed quantity---and is blind to the $L-T$ \emph{gap}, it cannot inflate the Text-vs-Latent
comparison. Both variants for all games are reported in Appendix~\ref{app:full}. The robustness 
choice itself is informative: it \emph{rescues} games whose bare action head collapses under 
suffix/bridge inputs (River~Raid $L\!:\!360\!\to\!612$, Q*bert $0\!\to\!50$), but \emph{harms} 
games where the bare head already works well (MsPacman $628\!\to\!60$, Seaquest $80\!\to\!0$). 
Even without any selection, the cross-game predictor (\S\ref{sec:predictor}) remains strong
($r=0.93$ on the 8 reported cells, $r=0.96$ on all 16 evaluated cells).

\paragraph{Games.} A total of nine Atari games were attempted. \emph{Frostbite} is excluded since action-head cloning converged to random val-accuracy, so no F/T/L comparison is meaningful. \emph{Pong} is reported for completeness but provides little insight: its action-head validation accuracy reaches only 25.1\% (versus 16.7\% random, or $1.5\times$ baseline). Because the bare Fast-Only ($F$) policy cannot reliably move the paddle, no bridge is able to rescue its performance. The remaining seven games are MsPacman, Seaquest, SpaceInvaders, RoadRunner, River~Raid, Enduro, Q*bert (shown with MetaDrive in Figure~\ref{fig:games}).

\section{Architecture: v1 cross-attention failed; v2 LLaVA-style works}
\label{sec:arch}

\begin{figure}[t]
\centering
\includegraphics[width=\textwidth]{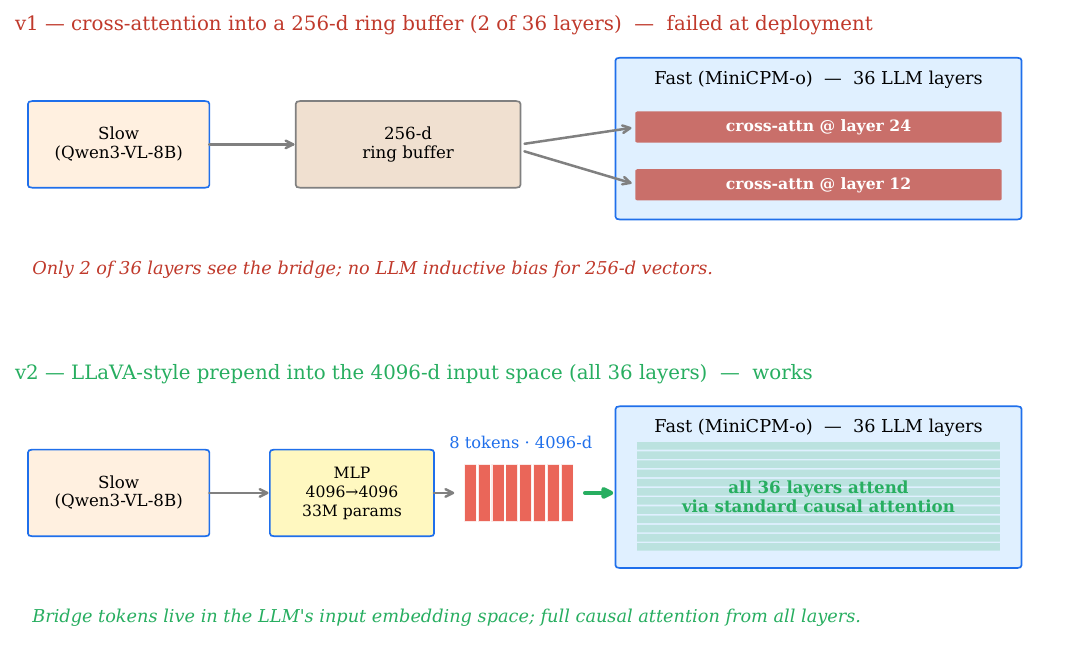}
\caption{\textbf{v1 vs v2 bridge architectures.} v1 (top, failed): 256-d cross-attention at 2 of 36 layers. v2 (bottom, working): 4096-d prepend, all 36 layers attend.}
\label{fig:arch}
\end{figure}

We tried two bridge architectures (Figure~\ref{fig:arch}). v1 projected slow-model residuals into a 256-d ring buffer and injected via cross-attention at LLM layers 12 and 24 (2 of 36) of the fast model. \emph{Despite converging to KL=0.004 on offline data}, v1 failed at deployment: the Latent Bridge scored 225 on MsPacman against Fast-Only's 256, bimodal with 4/12 catastrophic episodes. The three v1 variants (ungated, gated, gated+head-tune) all failed.

\paragraph{v2.} We adopt the same pattern LLaVA~\cite{llava}, BLIP-2~\cite{blip2}, Flamingo~\cite{flamingo}, and MiniCPM-o~\cite{minicpmo} use to couple vision and language: project slow residuals into the fast LLM's input embedding space (4096-d) via a 33\,M-parameter 2-layer MLP, and prepend the resulting 8 tokens to the fast model's input sequence. All 36 LLM layers attend over them via standard causal attention. Only the projection trains; both base models are frozen. The one firm methodological lesson is that \emph{offline KL convergence is necessary but not sufficient}---v1 converged and still lost to Fast-Only---so adapter-style fast/slow coupling should be validated at deployment, not by offline KL alone.

\section{Results}
\label{sec:results}

\paragraph{Evaluation protocol.} We evaluate each (game, strategy, variant) combination using 3 random seeds $\times$ 4 episodes per seed, for a total of $n=12$ episodes per cell. Episodes are capped at 500 ticks. Arcade Learning Environment (ALE) seeds vary per episode. Action selection is greedy (the action with the highest probability) except where we tune the decoder per channel for the headline (\S\ref{sec:best_achievable}); under greedy, many cells exhibit zero per-episode variance (Appendix~\ref{app:stats}).

\subsection{Headline: the Latent Bridge is a safe-or-better drop-in for the Text Bridge}
\label{sec:headline}

\begin{table}[h]
\centering\small
\input{figures/best_achievable.tex}
\caption{\textbf{Headline: best-achievable F/T/L per game}, each channel at its own action decoder
selected on held-out seeds (leave-one-seed-out; $n\!=\!12$; modal decoder in parentheses).
Significance is by Mann--Whitney $U$, which is appropriate because the best-achievable cells are non-normal
(MsPacman's $L$ is bimodal); Welch's $t$ agrees on RoadRunner and is borderline on MsPacman
($p\!=\!0.06$). Reported action-head variant per game (\S\ref{sec:setup_variant}); both variants in
Appendix~\ref{app:full}.}
\label{tab:headline}\label{tab:best_achievable}
\end{table}

\begin{figure}[t]
\centering
\includegraphics[width=\textwidth]{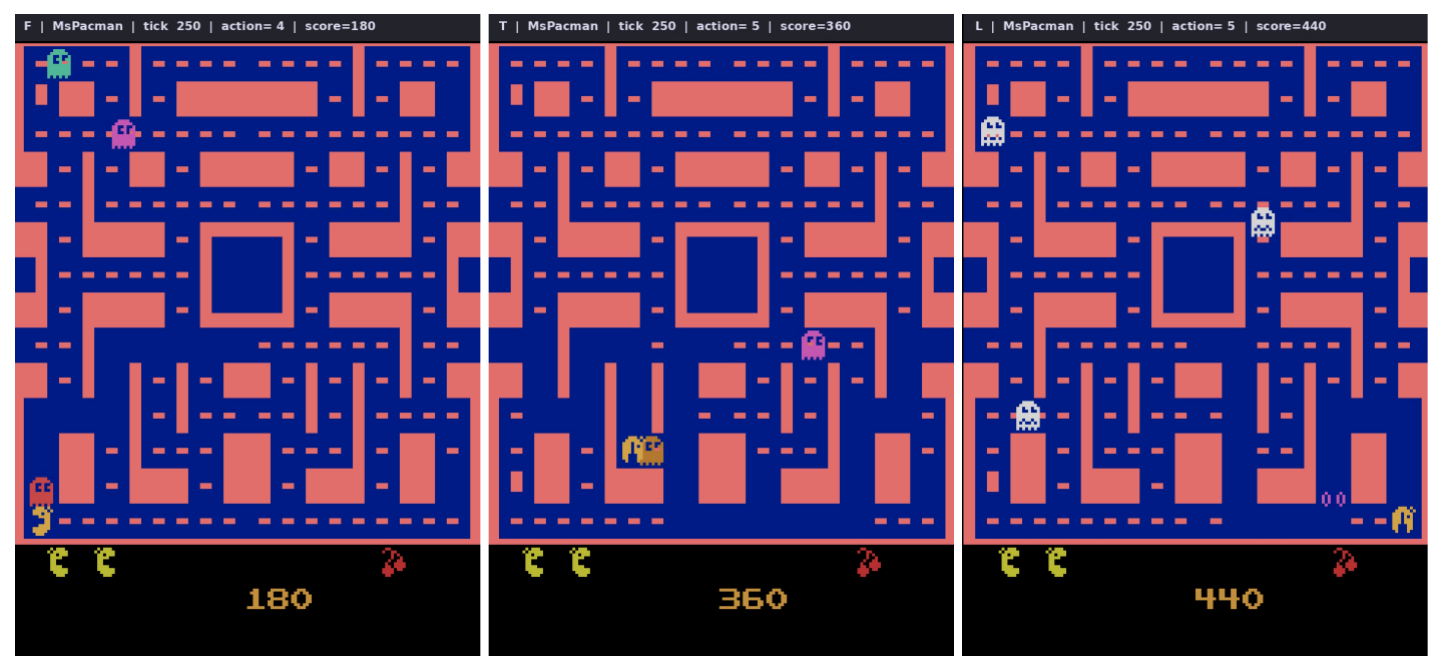}
\caption{\textbf{Ms.\ Pac-Man, one representative episode (same tick across channels).}
Fast-Only ($F$, left) receives no slow guidance; the Text Bridge ($T$, middle) and Latent Bridge
($L$, right) act on the slow model's emission and here lead $F$ on score ($180/360/440$). This
single frame is illustrative only: the headline $+57\%$ of $L$ over $T$ is an episode-mean over
$n\!=\!12$ (Table~\ref{tab:headline}), and Ms.\ Pac-Man's per-episode $L$ is bimodal, so the frame
is not a substitute for the pooled statistics.}
\label{fig:mspacman}
\end{figure}

We compare each channel using its own best action decoder, selected on held-out seeds (why this is the
fairer comparison, and how it differs from the rosier fixed-greedy number, is \S\ref{sec:best_achievable}).
Under this protocol (Figure~\ref{fig:headline}, Table~\ref{tab:headline}) the Latent Bridge is
\textbf{never significantly worse} than the Text Bridge and \textbf{significantly better on 2 of 7
games}: MsPacman ($+57\%$, Figure~\ref{fig:mspacman}) and RoadRunner ($+28\%$, the cleanest case, Figure~\ref{fig:roadrunner}).
The other five are statistical ties. So at matched architecture, the Latent Bridge is a
safe-or-better drop-in for the Text Bridge wherever the slow model is worth coupling at all.

\paragraph{The bridges do not always beat Fast-Only.}
On River~Raid and SpaceInvaders, Fast-Only outscores both bridges. While the Latent Bridge improves over
the Text Bridge \emph{at matched architecture}, whether the slow--fast architecture beats Fast-Only
at all is a separate, game-dependent question---and the predictor of \S\ref{sec:predictor} says when.

\subsection{RoadRunner: the cleanest demonstration}

\begin{figure}[t]
\centering
\includegraphics[width=0.65\textwidth]{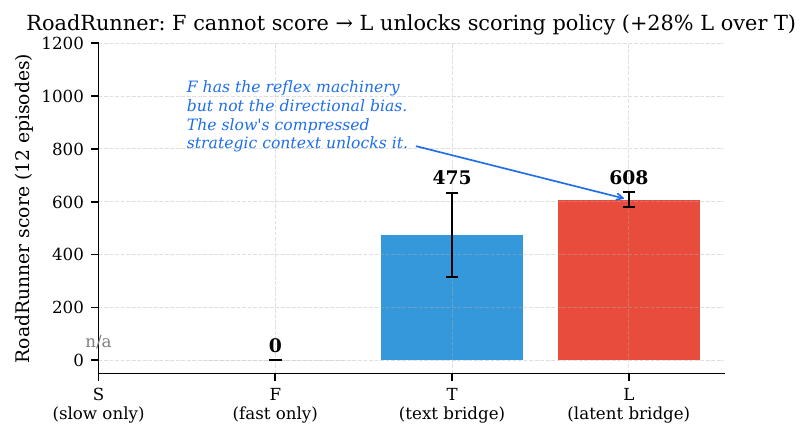}
\caption{\textbf{RoadRunner: F cannot score; L unlocks the policy.} Under bare action head, the action head's most-confident output is no-op even when the Coyote is closing. The slow's guidance (``head right'') breaks the local maximum. $L=608$ vs $T=475$, +28\%.}
\label{fig:roadrunner}
\end{figure}

RoadRunner is the cleanest qualitative demonstration (Figure~\ref{fig:roadrunner}): under bare action head the action head cannot score at all (its most-confident move is no-op, even as the Coyote closes), and the slow model's full joint-state emission (Coyote $\Delta_x$, obstacle and pellet coordinates) unlocks a scoring policy---with the Latent Bridge scoring higher than the Text Bridge. Two caveats keep this honest. The zero-score floor is specific to the \emph{bare} head---a robust head scores well and all three strategies tie (\S\ref{sec:ood})---so this shows the bridge rescuing a brittle baseline, not a fundamental planning limit. Also, because the baseline is exactly zero, the absolute magnitude is unstable run-to-run even though the $L\!>\!T$ direction is robust (Appendix~\ref{app:decoder_sweep}). Direct inspection (\S\ref{sec:bridge_content}) shows the Latent Bridge encodes mostly game identity plus a thin per-emission strategic residual---non-lexical but game-conditioned.

\subsection{Greedy decoding flips deterministic cells either way}
\label{sec:qbert}

Some cells are deterministic under greedy decoding: with a small action space and a confident
head, all 12 episodes follow the same trajectory, so the cell has zero variance and the
Latent-vs-Text comparison reduces to comparing two integers. On such cells the greedy sign is an
artifact and it cuts \emph{both ways}: greedy hands the Text Bridge a lead on Q*bert and the
Latent Bridge a lead on Seaquest, and \emph{both} examples reverse to ties once actions are sampled (full
per-temperature numbers in Appendix~\ref{app:decoder},~\ref{app:other_sampling}).

\paragraph{Why.} The Latent Bridge's per-emission variation is tiny (3--8\%,
\S\ref{sec:bridge_content})---below the threshold needed to flip a greedy argmax, so every episode
repeats. However, it is enough to move a \emph{sampled} action distribution. Which channel sampling
favors is game-specific, not a law. The takeaway is that greedy determinism makes the
latent-vs-text \emph{sign} unreliable on these zero-variance cells---which is exactly why the
headline tunes the decoder per channel rather than trusting a single fixed one
(\S\ref{sec:best_achievable}).

\subsection{Why best-achievable, not a fixed decoder: greedy would mislead}
\label{sec:best_achievable}

Why tune the decoder per channel rather than fix one for both channels? Because a single fixed decoder is unfair to
both channels at once---they peak at \emph{different} decoders (the Latent Bridge at greedy or
high-$\tau$, the Text Bridge at $\tau\!\approx\!0.5$). A fixed \emph{greedy} decoder in particular
would tell a rosier story: the Latent Bridge would appear to win \emph{four} games rather than two.
But those two extra wins are greedy-specific. They vanish at every fixed sampling temperature (full
sweep, Appendix~\ref{app:decoder_sweep})---the Latent Bridge's $3$--$8\%$ per-emission variance
(\S\ref{sec:bridge_content}) nudges the deterministic argmax but is swamped the moment the decoder
samples---and on zero-variance cells the greedy sign is an artifact that flips \emph{either} way
(it inflates the Text Bridge on Q*bert and the Latent Bridge on Seaquest; \S\ref{sec:qbert}). The
held-out selection is also unbiased rather than optimistic: it can land \emph{below} a channel's
in-sample greedy score, so the reported $+57\%$ on MsPacman is conservative, not inflated.

\paragraph{Held-out selection.} The fair comparison gives each channel its own best decoder, chosen
the honest way: tune on held-in seeds, report on held-out. With 3 ALE seeds (4 episodes each) we use
leave-one-seed-out---for each held-out seed, pick the decoder with the best mean on the other two,
then score the held-out seed under it---and pool the three folds (details in
Appendix~\ref{app:heldout}). This produces the headline of \S\ref{sec:headline}
(Table~\ref{tab:headline}): it shrinks the claim from four greedy wins to two decoder-robust ones,
but makes it honest. The task-level predictor (\S\ref{sec:predictor}) is unaffected---it compares
each bridge to Fast-Only, and the best bridge beats the best Fast-Only exactly where slow reasoning
helps ($T\!>\!F$).

\subsection{Using both channels at once hurts: the channels interfere, not complement}
\label{sec:combined}

A natural follow-up to ``the Latent and Text channels prefer different decoders and win on different
games'' is: \emph{use both at once.} We added a combined strategy $B$ that puts the slow model's text
suffix in the prompt \emph{and} prepends its latent tokens in the same forward pass, and ran it across
all 7 games at the same per-game decoder grid, with the same held-out best-achievable selection
(Appendix~\ref{app:heldout}). Table~\ref{tab:combined} compares best-$B$ to the best \emph{single}
channel per game.

\begin{table}[h]
\centering\small
\input{figures/combined.tex}
\caption{\textbf{Combining both channels (best $B$) vs.\ the best single channel}, held-out
best-achievable, $n\!=\!12$. $\Delta_{B-\text{single}}$ is $(B-\text{best single})/\text{best single}$.}
\label{tab:combined}
\end{table}

\paragraph{The naive ``more context is better'' intuition fails.} Combining the channels never
significantly beats the better single channel and significantly \emph{hurts} on 3 of 7 games
(catastrophically on RoadRunner; Table~\ref{tab:combined}). The damage concentrates exactly where
one channel is strong on its own---both the latent-dominated and the text-dominated games degrade
when the second signal is added. We attribute this to the frozen action head, which is trained on a
\emph{single} conditioning signal at a time (action-head cloning sees a bare prompt or a suffix; the bridge
is distilled toward the Text Bridge), so when it is handed both at once, it produces a worse policy.
\textbf{The design rule is therefore: couple via exactly one channel.} The predictor
(\S\ref{sec:predictor}) says \emph{whether} to couple ($T\!>\!F$); if so, pick a single channel,
with the Latent Bridge as the safe-or-better default.

\section{Mechanism: the action head is brittle to out-of-distribution inputs}
\label{sec:ood}

\begin{figure}[t]
\centering
\includegraphics[width=0.78\textwidth]{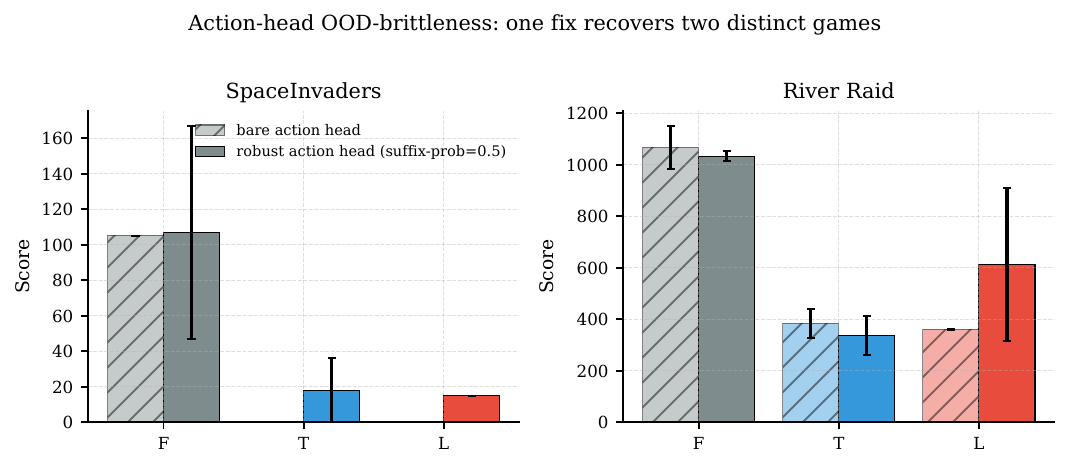}
\caption{\textbf{One action-head retraining recovers two distinct collapse modes.} SpaceInvaders (left): bare collapses both bridges to zero; robust lifts them off the floor (though still below Fast-Only). River~Raid (right): bare is a near-tie far below Fast-Only; robust leaves the Text Bridge flat but lifts the Latent Bridge well above it under greedy decoding---a margin that itself becomes a tie once the decoder is tuned per channel (\S\ref{sec:best_achievable}), which is why River~Raid is a tie in the headline. Greedy scores in Appendix~\ref{app:full}.}
\label{fig:ood}
\end{figure}

Three games went wrong in their bare-action-head sweeps despite cleanly converging bridge distillation: SpaceInvaders and Q*bert collapsed to zero, and River~Raid fell to a bridge-harmful near-tie far below Fast-Only. The same robust-action-head retraining recovers two distinct collapse modes (Figure~\ref{fig:ood}). The collapse is not a bridge-training failure: on SpaceInvaders, three independent interventions (random-policy and expert-policy $T$-trajectories, and an aggressive ``FIRE constantly'' slow prompt) all left it intact.

\paragraph{Diagnosis: a direct measurement of the drift.} Action-head cloning trains the action head on the \emph{bare} game-state prompt; at deployment, the text suffix and the bridge tokens are both OOD inputs. We measure the effect directly (Appendix~\ref{app:ood_kl}): on bare action head, the suffix \textbf{flips the action head's argmax on most emissions (70--90\%) of every game}---the drift is large and uniform. So the perturbation is real; what differs across games is \emph{what action the drift moves toward}---on some games the new argmax still scores, on others it lands on a bad action and scoring collapses, a property of each game's reward structure rather than of the bridge. The bridge mechanism is fine; the frozen action head's suffix-induced argmax flip is the real bottleneck.

\paragraph{The fix.} Retrain the action head at suffix-probability 0.5: half of training batches receive a randomly-chosen slow-style text suffix appended to the prompt. The head learns suffix-invariance.

\paragraph{The fix is not universal.} The robust action head rescues the games whose bare bridges collapsed (SpaceInvaders, River~Raid---dramatically lifting the Latent Bridge on River~Raid) but \emph{harms} the games whose bare bridges already worked, destroying the Latent Bridge on MsPacman and Seaquest and erasing the advantage on RoadRunner (where it instead lifts Fast-Only off its zero floor, so all three strategies tie). \textbf{Recommendation}: use the robust action head only when a bare bridge collapses to near zero; otherwise use the bare one. Per-game scores are in Appendix~\ref{app:full}.

\paragraph{The robust action head reduces drift exactly where expected.} On the non-collapsed games, robust
action head sharply cuts the suffix-induced argmax drift (e.g.\ RoadRunner $75\%\!\to\!12\%$)---except
Seaquest, whose drift \emph{rises}, matching robust action head's deployment harm there. On the collapsed
games it lowers the KL but only partly reduces the drift: the head grows more confident under the
suffix without always landing on a recoverable action. The full per-game probe is in
Appendix~\ref{app:ood_kl}.

\section{Channel ablations}
\label{sec:ablations}

\subsection{The channel is not bandwidth-limited}
\label{sec:bandwidth}

Sweeping the latent token count $N\!\in\!\{4,8,16\}$ on MsPacman shows the channel does not behave like a
bandwidth bottleneck. Trained and deployed at a matched $N$, the Latent Bridge peaks at
$N\!=\!8$ ($628$, vs $296$/$259$ at $N\!=\!4$/$16$); but \emph{deploying} a bridge trained at
$N\!=\!8$ with more positions ($N\!=\!16$) scores higher still ($720$), exploiting positions it
was never trained for. The gating quantity is thus not how many tokens the channel carries---%
consistent with the low per-emission variance of \S\ref{sec:bridge_content}---so we do not
advance a ``text is bandwidth-limited'' account.

\subsection{A longer text suffix does not close the gap---it widens it}
\label{sec:longer_t}

The simplest objection to the Latent Bridge's advantage is ``the Text Bridge just needs more
text.'' We tested it directly: on MsPacman and RoadRunner we re-ran the Text Bridge with a
rolling window of the last $w\!\in\!\{1,2,3\}$ slow emissions concatenated into the suffix,
holding the model, checkpoint, seeds, and episode count fixed (Figure~\ref{fig:longer_t}---full
numbers in Appendix~\ref{app:longer_t}). Concatenating older emissions \emph{hurts}: stale
emissions describe game state that no longer holds, and the action head cannot disambiguate their
time ordering. The decline is gentle on MsPacman (slower-changing pellet strategy) and sharp on
RoadRunner (moment-to-moment obstacle dodging), where every episode scores zero by $w\!=\!3$.
Because the Latent Bridge transmits only the most-recent emission, its score is unchanged by $w$,
so the gap only widens. To compete, the Text Bridge would need explicit time-tagging or a learned
attention over emissions.

\begin{figure}[t]
\centering
\includegraphics[width=\textwidth]{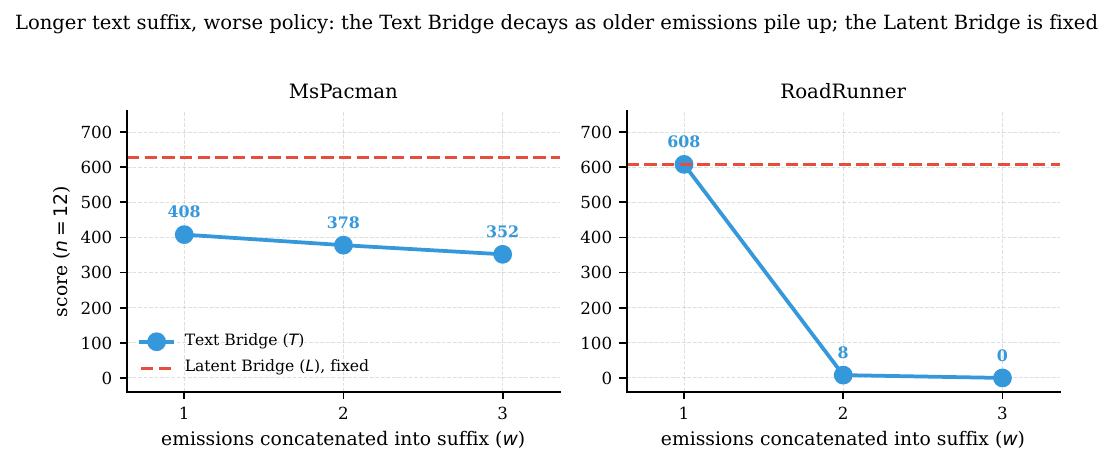}
\caption{\textbf{A longer text suffix widens the gap.} The Text Bridge's score as a function of
how many recent slow emissions are concatenated into the suffix; the Latent Bridge (latest
emission only) is the fixed dashed line. (RoadRunner is run-to-run-variable, \S\ref{sec:results};
shown is one internally consistent run, where the within-run $w\!=\!1$ baseline starts at 608 and
collapses---the trend, not the headline level.)}
\label{fig:longer_t}
\end{figure}

\subsection{The bridge tokens carry real signal, not just extra positions}
\label{sec:bridge_replace}

A negative-control objection to the Latent Bridge: maybe the trained tokens do nothing, and the
LLM merely benefits from 8 extra prepended positions to compute over. We tested it on MsPacman by
replacing the trained projection's output with 8 zero vectors or 8 random vectors at the trained
per-position norm, holding everything else fixed (Figure~\ref{fig:bridge_decomp}; significance
tests are in Appendix~\ref{app:bridge_replace}). Both effects are real and separable: the empty slots
alone lift the policy $\approx$40\% over Fast-Only (a pause-token-like effect~\cite{pause_tokens}),
and the \emph{trained} content adds the remaining $\approx$60\%---the part that distinguishes the
Latent Bridge from a generic ``more compute per tick'' baseline. This single-game split is not
universal: the all-games version of this control (\S\ref{sec:learned_content}) finds the learned
fraction ranges from $\approx$100\% (RoadRunner) to negative (River~Raid), with its sign predicted
by whether slow reasoning helps the task ($T\!>\!F$).

\begin{figure}[t]
\centering
\includegraphics[width=0.74\textwidth]{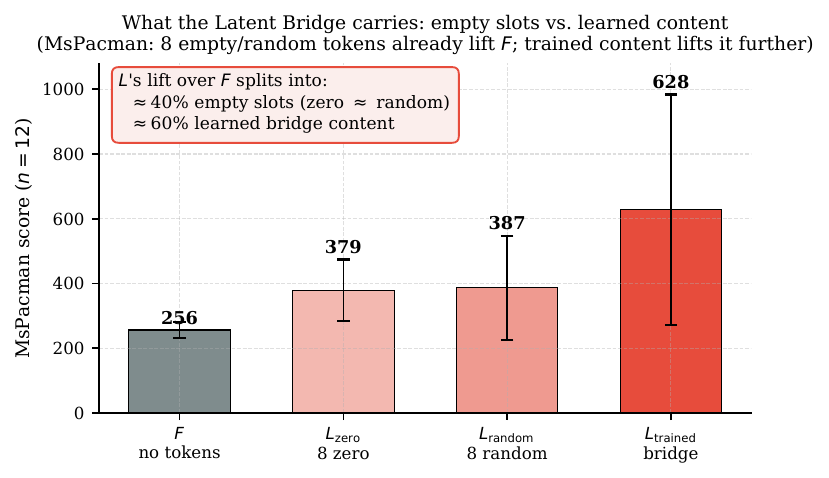}
\caption{\textbf{The latent's lift over Fast-Only splits into architectural ``slots'' and learned
content (MsPacman).} Eight zero or random prepended tokens already supply the ``slots'' part; the
trained bridge adds the rest. Zero and random are statistically indistinguishable
(Appendix~\ref{app:bridge_replace}).}
\label{fig:bridge_decomp}
\end{figure}

\subsection{What the bridge encodes: mostly game identity, not text}
\label{sec:bridge_content}

We project the trained bridge tokens of each game (with cached slow residuals from seed-0 trajectories, 8 emissions per game) into the fast LLM's input embedding space and compare them directly to the fast LLM's 151{,}748 vocab embeddings.

\paragraph{The bridge is non-lexical.} Bridge tokens have L2 norm $\sim$64 (constrained by the output LayerNorm); vocab embeddings have norm $\sim$1.45 (p99 1.92). The cosine similarity of any bridge token to its nearest vocab embedding is $\le 0.09$ across all 7 games. Decoding each bridge token to its nearest vocabulary embedding returns visually random tokens (rare Unicode characters, code-snippet fragments), with similarities no better than chance---at the 99.9th percentile of all vocabulary similarities. \textbf{The latent channel does not write its content in the space of word embeddings}---it lives in a region of $\mathbb{R}^{4096}$ that the frozen LLM has learned to attend to but that no text token occupies. This is consistent with what LLaVA's vision projection produces (also non-lexical). Thus, the right answer to the question ``can we just read the latents as text?'' is we cannot.

\paragraph{The bridge is game-conditioned.} Bridge tokens of the \emph{same} game cluster tightly (mean pairwise cosine \textbf{+0.80 to +0.89}) while \emph{different} games are nearly orthogonal (\textbf{+0.04 to +0.25}), with related games (the spatial-2D set) closer than unrelated ones (forward-driving Enduro). The full $7\!\times\!7$ matrix is in Appendix~\ref{app:bridge_diag}.

\paragraph{The bridge is weakly per-emission-conditioned.}
Cosine between the \emph{mean bridge token} of one emission vs another emission \emph{within the same game}: +0.92 to +0.97 (std 0.02--0.03). The bridge varies only $\sim$3--8\% between consecutive emissions of the same game. This is a very low per-emission variance and explains the deterministic-trajectory outcomes (e.g., Q*bert's std=0 cells): if every emission produces near-identical bridge tokens, the fast model picks near-identical actions, so 12 episodes converge on the same trajectory under greedy decoding.

\paragraph{Implication: the bridge uses far less capacity than it has.} The latent channel's nominal capacity is enormous, but its \emph{used} capacity is small: most of its variance encodes the (episode-fixed) game identity, and only a thin fraction encodes per-emission strategic state. So the latent's advantage comes from using only part of its capacity, not from running into a ceiling. This is why what gates the bridge is whether the slow channel carries a useful signal at all (\S\ref{sec:predictor}), not how many bits it could carry.

\section{When the bridge helps: a behavioral predictor}
\label{sec:predictor}

The headline tally (\S\ref{sec:results}) says \emph{where} the Latent Bridge wins, and the
ablations (\S\ref{sec:ablations}) say the reason is not bandwidth; this section asks
\emph{when} the bridge pays off. Across every domain we evaluated, a single behavioral
variable---whether slow reasoning helps the task at all ($T\!>\!F$)---predicts it. We first
extend the testbed beyond Atari (\S\ref{sec:metadrive}), then state the predictor
(\S\ref{sec:predictor_stmt}) and back it up with a mechanistic check (\S\ref{sec:learned_content}),
and close with an axis that does \emph{not} predict when the bridge pays off(\S\ref{sec:no_lexical}).

\subsection{Non-Atari test: a driving domain}
\label{sec:metadrive}

To probe whether the bridge generalizes beyond Atari, we ported the full pipeline to
\textbf{MetaDrive}~\cite{metadrive}, a real-time driving simulator with the same fast/slow
premise (a $\sim$10\,Hz reactive control loop; a $\sim$1.5\,s reasoning step that cannot sit
in the loop). We use a top-down rendered observation for the fast model and the same structured
state decoder (speed, lane offset, heading, upcoming-turn cue, neighbours) for the slow model.
This is, to our knowledge, the first non-game test of the Latent Bridge; the integration runs
end-to-end on one GPU (Appendix~\ref{app:metadrive}).

\begin{figure}[t]
\centering
\includegraphics[width=\textwidth]{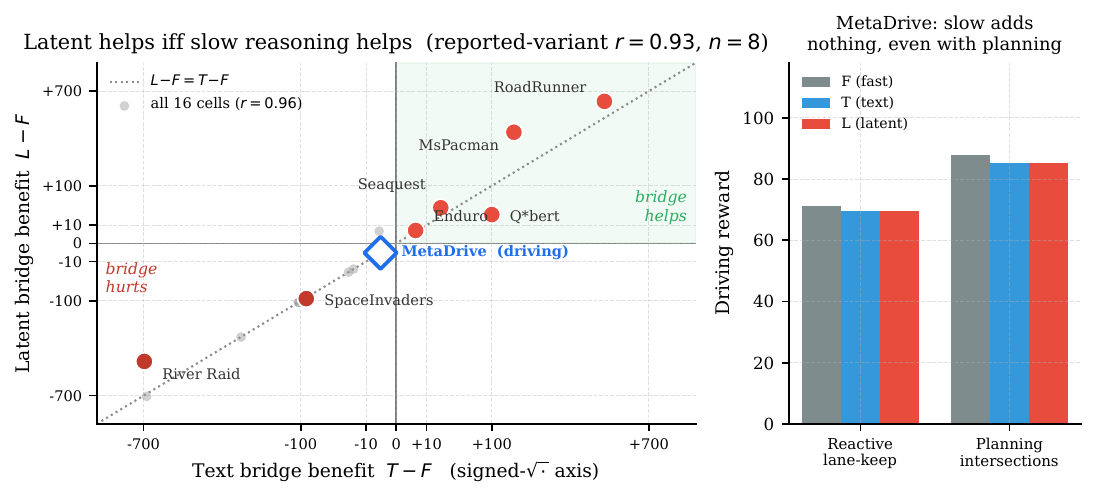}
\caption{\textbf{The Latent Bridge helps if and only if the slow model helps with the task.}
\emph{Left:} per-game Latent benefit $L\!-\!F$ vs.\ Text benefit $T\!-\!F$ across 7 Atari games
and MetaDrive, on a signed-$\sqrt{\cdot}$ axis. Bold points are the reported-variant cell per game
(Pearson $r=0.93$, $n=8$); faint grey points are all 16 evaluated cells ($r=0.96$), so the
relationship is not an artifact of variant selection (robustness stats in
Appendix~\ref{app:metadrive}). The bridge helps only in the upper-right quadrant ($T\!>\!F$ and
$L\!>\!F$); MetaDrive (blue diamond) sits at the origin. \emph{Right:} in both MetaDrive
regimes---reactive lane-keeping and planning-heavy intersections---slow reasoning never exceeds
Fast-Only, even when the task demands route planning.}
\label{fig:predictor}
\end{figure}

\paragraph{The bridge does not transfer to driving---and the controls explain why.}
On the default lane-keeping map the Latent Bridge ties with Fast-Only and the Text Bridge collapses. The
bridge-replacement control (Appendix~\ref{app:bridge_replace}, method) shows the Latent Bridge is \emph{inert}
(zeroing or randomizing the bridge tokens leaves the score unchanged, $L\!\approx\!L_{\text{zero}}\!\approx\!L_{\text{random}}\!\approx\!F$).
This is \emph{not} the action-head OOD artifact of \S\ref{sec:ood}: it persists on the
robust head, under sampling, and after redistilling the bridge on expert-driven trajectories with a
healthy ($T\!\approx\!F$) teacher.

\paragraph{The failure is not that the task lacks a planning component.}
We rebuilt MetaDrive as a \emph{planning-heavy} route (straights punctuated by intersections,
with off-route termination) where survival requires deliberate turn decisions---the
navigation expert beats a naive straight-driving policy by $+118$ reward, and the slow model
receives explicit turn cues. Even here (Figure~\ref{fig:predictor}, right) the slow model does not
beat Fast-Only---it ties under greedy and actively \emph{hurts} under sampling, with the Latent
Bridge tracking the Text Bridge down (numbers in Appendix~\ref{app:metadrive}). Driving is dominated
by a tight visuomotor loop the fast model already handles; the slow model's reasoning---slow to
arrive and tied to a single past frame---does not improve on it, unlike Atari planning games whose
slower dynamics reward deliberation.

\subsection{The predictor: the Latent Bridge helps iff the slow model helps}
\label{sec:predictor_stmt}

That the Latent Bridge tracks the Text Bridge is partly built in by construction: bridge
distillation trains the Latent Bridge to imitate the
Text Bridge's action distribution (minimizing $\mathrm{KL}(\pi_L\,\|\,\pi_T)$), so the Latent Bridge
can at best convey what the Text Bridge already conveys---it can only help where the Text Bridge
helps. The data bear this out. Pooling all eight game-evals (7 Atari + MetaDrive;
Figure~\ref{fig:predictor}, left), the Latent Bridge's benefit over Fast-Only ($L\!-\!F$) tracks
the Text Bridge's benefit ($T\!-\!F$) at \textbf{Pearson $r=0.93$} (bootstrap 95\,\% CI
$[0.81,1.0]$). The relationship is not an artifact of variant selection: it holds at $r=0.96$ over
all 16 evaluated cells---the 14 bare/robust Atari cells plus a SpaceInvaders expert-data cell and
MetaDrive, with no per-game selection (\S\ref{sec:setup_variant}, Appendix~\ref{app:metadrive}). The bridge pays off exactly
when slow reasoning beats reaction on the task ($T\!>\!F$: RoadRunner, MsPacman, Seaquest,
Q*bert; Enduro sits at floor, with $T\!-\!F$ a few points either side of zero depending on
variant) and is inert or harmful when it does not (River~Raid, SpaceInvaders, MetaDrive). This
reframes ``does a Latent Bridge help?'' as a property of the \emph{task} (is the bottleneck
deliberation or perception--action?), not of the channel---and makes MetaDrive a clean
controlled negative rather than a failed port.

\subsection{The bridge carries learned content only where slow reasoning helps}
\label{sec:learned_content}

The same bridge-replacement control (\S\ref{sec:bridge_replace}, worked through there for MsPacman)
also tells us, game by game, whether the Latent Bridge carries learned, behavior-relevant content.
We ran it on all 7 Atari games; ``learned'' means
$L_{\text{trained}}$ exceeds $\max(L_{\text{zero}},L_{\text{random}})$ by $>\!10\%$. Cells use
the bare action head under greedy decoding (Q*bert: robust head under $\tau\!=\!1$ sampling,
its canonical decoder); $T\!-\!F$ is the matching bare-head value, and the run is an independent
re-run whose repeated cells agree with the headline within run-to-run noise (Table~\ref{tab:learned_content};
full numbers and caveats in Appendix~\ref{app:bridge_replace}).

\begin{table}[h]
\centering\small
\begin{tabular}{lrrrll}
\toprule
Game & $L_{\text{train}}$ & $L_{\text{zero}}$ & $L_{\text{rand}}$ & verdict & $T\!-\!F$ \\
\midrule
RoadRunner    & $\mathbf{608}$ & 0     & 8     & learned & $+475$ \\
Seaquest      & $\mathbf{100}$ & 28    & 5     & learned & $+22$  \\
MsPacman      & $\mathbf{666}$ & 408   & 410   & learned & $+152$ \\
Enduro        & 7.8            & 1.4   & 4.7   & trend only (at floor) & $-3$ \\
Q*bert        & 123            & 63    & 117   & $\approx$random & $+100$ \\
SpaceInvaders & 0              & 148   & 90    & \emph{harmful} & $-105$ \\
River Raid    & 360            & 1013  & 1003  & \emph{harmful} & $-683$ \\
\bottomrule
\end{tabular}
\caption{\textbf{Bridge-replacement control on all 7 Atari games.} The Latent Bridge score with the
trained projection ($L_{\text{train}}$) against the projection replaced by zeros ($L_{\text{zero}}$)
or by random vectors at matched norm ($L_{\text{rand}}$); $T\!-\!F$ is the matching predictor value.}
\label{tab:learned_content}
\end{table}

The verdict follows the predictor. The trained latent is learned exactly on the games
where slow reasoning helps ($T\!>\!F$: RoadRunner, Seaquest, MsPacman)---on RoadRunner the
controls drop to the floor, so the Latent Bridge there is almost entirely learned content. Where slow
reasoning does not help ($T\!\le\!F$: River~Raid, SpaceInvaders), the trained latent is
\emph{harmful}: zeroing or randomizing it scores \emph{better}---the same inert/harmful pattern
as MetaDrive (\S\ref{sec:metadrive}). The two off-diagonal cases are benign: Enduro's scores are
too small for the comparison to be meaningful, and Q*bert---the decoder-fragile game
(\S\ref{sec:qbert})---shows trained\,$\approx$\,random under sampling, consistent with its thin
per-emission signal. The latent helps where, and only where, it has carried real content from a
slow channel that itself beats Fast-Only.

\subsection{What does not predict: emission statistics}
\label{sec:no_lexical}

\begin{figure}[t]
\centering
\includegraphics[width=0.78\textwidth]{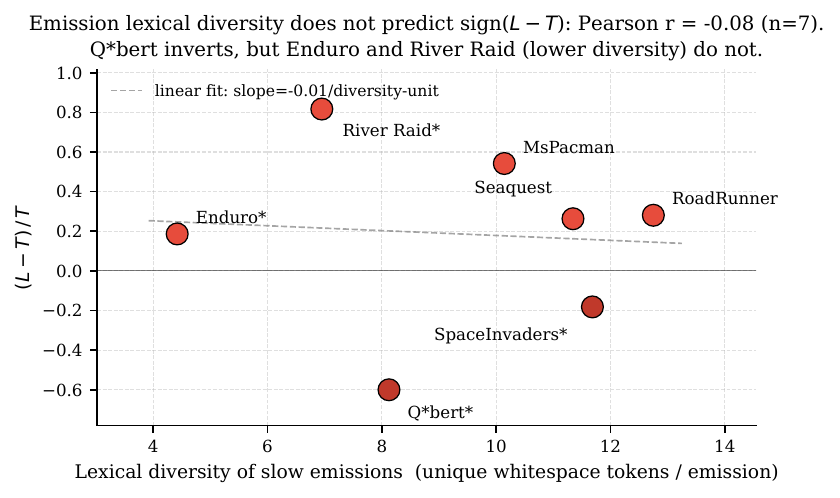}
\caption{\textbf{Emission statistics do not predict the sign of $L-T$ ($n=7$).}
$x$ = mean unique whitespace tokens per slow emission (seed-0 $T$-trajectory). $y = (L-T)/T$.
Q*bert is the only $y<0$ point with a large $T$, but Enduro and River~Raid (lower diversity) have $y>0$, and SpaceInvaders (high diversity) is mildly $y<0$. Pearson $r=-0.08$. The dashed line is the linear fit and is shown only to make the lack of structure visible.}
\label{fig:scatter}
\end{figure}

We also looked for a signal we could read off the emissions ahead of time. We guessed that the
lexical diversity of slow emissions (unique whitespace-split tokens per emission) would predict the
sign of $L\!-\!T$. Across our
7 games (Figure~\ref{fig:scatter}), $r=-0.08$ (slope $-0.01$ per diversity-unit; not
significant). Six alternative features (gzip ratio, numeric density, distinct numbers, token
counts) all top out at $|r|\!\le\!0.25$. We retain the figure as a negative result: emission
\emph{statistics} do not predict the sign of $L\!-\!T$.

\section{Discussion}
\label{sec:discussion}

\subsection{Limitations}

\begin{itemize}[topsep=0pt,itemsep=0pt,leftmargin=*]
\item \textbf{Small $n$.} We tested only 12 episodes per cell. Several cells have zero per-episode variance (greedy decoding + deterministic policy + small action space). On those, parametric statistics are not meaningful and we report MWU or just note the deterministic outcome.
\item \textbf{Per-game variant selection.} The headline reports, per game, the action-head variant with the higher deployment $L$ among variants whose $T$ has not collapsed (\S\ref{sec:setup_variant}); this is a per-game choice. We report both variants for every game in Appendix~\ref{app:full}, and show the cross-game predictor holds with no selection ($r=0.96$ over all 16 evaluated cells).
\item \textbf{The token-count ($N$) ablation is three points.} The deploy-only result (a bridge trained at $N\!=\!8$ scoring best deployed at $N\!=\!16$) is noted in \S\ref{sec:bandwidth}.
\item \textbf{Per-emission information content is not measured directly.} We show in \S\ref{sec:bridge_content} that the bridge is strongly game-conditioned but only weakly per-emission-conditioned; a quantitative bound on the Latent Bridge's per-emission information content is the natural next step.
\item \textbf{Beyond Atari: one driving domain, a controlled negative.} We ported the pipeline to MetaDrive (\S\ref{sec:metadrive}, Appendix~\ref{app:metadrive}). The bridge does not help there, and the bridge-replacement control shows why (the Latent Bridge is inert because the slow model's guidance never beats Fast-Only, $T\!\le\!F$). This is consistent with the $(T\!-\!F, L\!-\!F)$ predictor ($r=0.93$). However, this is a single non-game domain; we still have not tested the motivating target setting---real-time computer-use and game agents on phone and desktop, with their much larger action and observation spaces.
\end{itemize}

Two further negative results reinforce the picture: a longer rolling text suffix \emph{widens}
rather than closes the gap (\S\ref{sec:longer_t}), and swapping in a 30B-A3B MoE slow model does not
grow the Latent-over-Text gap on the one game tested (Appendix~\ref{app:scaling_30b}).

\subsection{Implications}

The practical rule is simple: measure the Text Bridge first. If slow reasoning does not beat fast reaction ($T\!\le\!F$, as on MetaDrive), neither bridge pays for itself---and the Text Bridge is the cheaper way to find that out. If it does ($T\!>\!F$), reach for the Latent Bridge: it is better by up to $+57\%$ where it wins and never significantly worse elsewhere.

\section{Related Work}
\label{sec:related}

\paragraph{Real-time interaction.} Real-time multimodal systems take several forms. Some handle understanding, generation, and interaction in a single model---Gemini Live~\cite{gemini_live}, OpenAI Realtime~\cite{openai_realtime}---and StepFun's open-weights \emph{Step-Audio-R1.1}~\cite{stepaudior1} folds explicit reasoning into one model by interleaving it with speech. Others decouple a fast responder from a slow reasoner that runs in the background---xAI's \emph{Grok Voice Think Fast}~\cite{grokvoice}, Thinking Machines Lab's \emph{Interaction Models}~\cite{interaction_models}, and Pine~AI's voice agent~\cite{pineai}. These are all tightly integrated, co-designed systems. We instead couple two \emph{independent, frozen, general-purpose} models and study only the coupling channel: the open VLMs one would deploy as a game or computer-use agent (MiniCPM-o~4.5~\cite{minicpmo}, Qwen3-VL~\cite{qwen3vl}) expose no such fast/slow split, and we test a \emph{learned latent} channel as the alternative to their text/context coupling.

\paragraph{Computer-use and GUI agents.} Operating a phone or desktop from pixels, these inherit exactly this tension between a tight perception--action loop and goal-level planning. SayCan~\cite{saycan} and Inner Monologue~\cite{inner_monologue} couple LLM planners to skill policies via natural-language sub-goals; RT-2~\cite{rt2} unifies them with a discrete-token action head. Our setting holds the architecture fixed and varies only the \emph{channel} between two frozen general-purpose multimodal LLMs at matched scale.

\paragraph{Multimodal coupling.} LLaVA~\cite{llava}, BLIP-2~\cite{blip2}, Flamingo~\cite{flamingo}, MiniCPM-o~\cite{minicpmo} couple vision and language by prepending projected visual tokens to an LLM's input. BLIP-2 ablates query-token count, the closest prior analog to our $N$ sweep. Our v2 transfers this pattern to language-language fast/slow coupling.

\paragraph{Continuous chain-of-thought.} COCONUT~\cite{coconut}, Pause Tokens~\cite{pause_tokens}, Implicit CoT~\cite{implicit_cot}, Quiet-STaR~\cite{quiet_star} all reason in latent space within a single model. Our work is the cross-model analog: two frozen models coupled by a learned latent channel.

\paragraph{Hierarchical / cascades.} Speculative decoding~\cite{speculative} couples draft+verifier within one decoding loop; Mixture-of-Depths~\cite{mod} adapts compute per token. These optimize a single forward pass; we decouple two models in wall-clock and pass the slower's deliberation forward as a learned signal.

\paragraph{Position.} To our knowledge this is the first reported head-to-head text-vs-latent comparison between two frozen general-purpose multimodal LLMs at matched scale, with deployment-time evaluation (not just offline KL) and a controlled negative domain (MetaDrive, where neither bridge helps) that bounds the claim.

\section{Conclusion}

We coupled two frozen, matched-scale multimodal LLMs (9\,B fast, 8\,B slow) through a learned
continuous Latent Bridge and compared it head-to-head with the Text Bridge fast/slow systems
normally use. Tuned per channel on held-out seeds---since the decoder is a deployment
hyperparameter---the Latent Bridge is a safe-or-better drop-in: never significantly worse than the
Text Bridge and significantly better on 2 of 7 games (\S\ref{sec:best_achievable}). Three results round out the
design rules: feeding \emph{both} channels to the frozen head does not combine their strengths but
instead \emph{interferes} (RoadRunner $-96\%$), so coupling should be done via exactly one (\S\ref{sec:combined}). Most
``bridge collapsed'' cells come from the action head being brittle to out-of-distribution inputs,
not from the bridge failing, and a targeted retraining recovers them (\S\ref{sec:ood}); and the
obvious objections do not hold---a longer text suffix \emph{widens} the gap (\S\ref{sec:longer_t}),
and replacing the trained latent with zeros or random vectors throws away most of the gain on the
games where it wins (\S\ref{sec:bridge_replace}).

Rather than a ``text is bandwidth-limited'' story---which our own ablations do not support
(\S\ref{sec:bandwidth}, Appendix~\ref{app:scaling_30b})---we land on a sharper claim:
\textbf{the Latent Bridge helps if and only if slow reasoning helps the task} ($L\!-\!F$ tracks
$T\!-\!F$ at $r=0.93$; MetaDrive is the controlled negative where the slow model never beats
Fast-Only and the Latent Bridge is correspondingly inert, \S\ref{sec:predictor}).

The cleanest follow-ups are (i) a training signal that rewards the bridge for telling consecutive
slow emissions apart, so more per-tick state lands in it---the trained Latent Bridge is non-lexical
and strongly game-conditioned but only weakly per-emission-conditioned, so it uses only a fraction
of its capacity (\S\ref{sec:bridge_content})---and (ii) scaling the comparison to the
target setting that motivates it: real-time computer-use and game agents on phone and desktop,
where the same fast-react/slow-plan tension holds but the action and observation spaces are far
larger than Atari.

\bibliographystyle{plain}
\bibliography{refs}

\appendix
\section{Full results table}
\label{app:full}

Both bare-action-head and robust-action-head reported for every game, $n=12$ per cell, under greedy
decoding; the headline (\S\ref{sec:headline}, Table~\ref{tab:headline}) instead uses the
best-achievable decoder selected per channel on held-out seeds (\S\ref{sec:best_achievable}).

\smallskip\noindent
\input{figures/stats_table.tex}
\smallskip

\paragraph{Reading the table.} Welch's $t$ becomes undefined or unreliable when either cell has zero variance; in those cases we report Mann--Whitney $U$ (marked $^{\dagger}$). On cells where \emph{both} sides have zero variance (e.g., Q*bert/robust $T=125\!\pm\!0$, $L=50\!\pm\!0$), the test reduces to comparing two integers; the $p$-value is mechanical, not a sampling statement.

\section{Statistics methodology}
\label{app:stats}

\paragraph{Variance sources.} Each cell is $3 \text{ seeds} \times 4 \text{ episodes per seed} = 12$ episodes. Seeds vary the ALE initial state. Action selection is greedy (argmax over logits). For games with small action spaces and policies that consistently pick the same action sequence, every episode within a seed often hits the same trajectory; hence many cells have $\sigma=0$. This is not a sampling artifact---it is the policy being deterministic given the env seed.

\paragraph{When we trust which test.}
\begin{itemize}[topsep=0pt,itemsep=0pt,leftmargin=*]
\item Both cells have $\sigma>0$ and roughly normal: \textbf{Welch's $t$}.
\item One cell has $\sigma=0$, the other has support: \textbf{Mann--Whitney $U$}.
\item Both cells have $\sigma=0$: \textbf{we report the comparison but do not claim a $p$-value} (the test is deterministic-trajectory difference, not a population inference).
\end{itemize}

\paragraph{Test convention.} All Welch's $t$ values use the unbiased sample-variance form, as in SciPy's \texttt{ttest\_ind} with \texttt{equal\_var=False}.

\paragraph{Bootstrap CIs.} $10^4$ resamples for each cell mean.

\paragraph{Cohen's $d$.} Reported only when both cells have $\sigma>0$. Zero-variance cells produce mechanically large $d$ that we do not interpret.

\section{Slow emission samples}
\label{app:emissions}

\paragraph{RoadRunner.}
\begin{quote}\small\textit{``Got it, let's break this down. The current state: Road Runner is at (3,0), Coyote is at (0,0) with a $\Delta_x$ of +3, so Coyote is closing in. The nearest birdseed is at $x=159$, which is ahead. There are obstacles at (170,16) and (170,0), so trucks/landmines are in the way. \dots''}\end{quote}

\paragraph{Q*bert.}
\begin{quote}\small\textit{``Got it, let's analyze the Q*bert state. The player is at (0,0), Coily is at (65,0) which is on the right edge. The purple and green balls are at $y=148$, so they're falling. The goal is to jump tiles to change colors, avoid Coily and balls. First, threat: Coily is on the right edge, so if Q*bert moves\dots''}\end{quote}

Both samples are $\sim$300 characters. RoadRunner's encodes simultaneous $\Delta_x$ for Coyote, target $x$ for birdseed, two obstacle coordinates, and a directional commitment. Q*bert's encodes (jump-direction, target-color, threat-actor)---a triple. This is the qualitative content that motivated the lexical-diversity hypothesis tested (and rejected) in \S\ref{sec:no_lexical}; we make no quantitative prediction on top.

\section{Action-head and bridge-distillation details}
\label{app:training}

\paragraph{Action-head cloning.} AdamW, lr $1\!\times\!10^{-4}$, 3 epochs, batch=1 with grad-accum=8. The action
head has 18 outputs (full ALE action space) on every game; the random baseline below is $1/k$ for
$k$ = the game's ALE \emph{minimal} action set (the only actions the SB3 expert emits, so the
val-set classes). Bare val-accuracies on the game-state-only prompt: MsPacman 32\% (random 11.1\%,
$k\!=\!9$), Seaquest 24\% (5.6\%, $k\!=\!18$), SpaceInvaders 33\% (16.7\%, $k\!=\!6$), Pong 25\%
(16.7\%, $k\!=\!6$), RoadRunner 59\% (5.6\%, $k\!=\!18$), River~Raid 31\% (5.6\%, $k\!=\!18$),
Enduro 49\% (11.1\%, $k\!=\!9$), Q*bert 34\% (16.7\%, $k\!=\!6$). The robust action head is trained identically except
half of training batches receive a synthetic slow-style text suffix appended (suffix-prob=0.5);
bare val-acc barely changes (e.g., MsPacman 32$\to$33\%, SI 33$\to$30\%) but the policy is
suffix-invariant.

\paragraph{Bridge distillation.} 2-layer MLP projection (LayerNorm, Linear 4096$\to$4096, GELU, Linear 4096$\to$4096, LayerNorm), 33.6\,M parameters. AdamW, lr $5\!\times\!10^{-5}$, 1 epoch, batch=1 with grad-accum=4. KL temperature $\tau=1.0$. Slow residuals extracted at layer 24 of 36 (we did not ablate the layer choice). Default $N=8$ latent tokens, $\sim$5K (frame, slow text, slow residuals) tuples per game from cached random-policy rollouts.

\paragraph{Text-bridge suffix.} The slow's full unmodified post-thinking emission, appended verbatim under a ``strategic-guidance'' tag; no truncation; median 302 chars across games.

\paragraph{Slow prompt template (excerpt, RoadRunner, paraphrased).}
The object coordinates fed to the slow model are extracted from the ALE RAM in the style of the
Atari Annotated RAM Interface~\cite{atariari}.
``Road Runner is at $(x_{rr}, y_{rr})$, Coyote at $(x_c, y_c)$ with $\Delta_x={+}3$. Nearest birdseed at $x=159$; obstacles at $(170,16)$ and $(170,0)$. \dots'' The system prompt instructs the slow to emit 1--3 sentences identifying threat/opportunity and recommended intent---\emph{not} individual actions.

\section{Latency breakdown}
\label{app:latency}

All experiments run on a single NVIDIA RTX Pro 6000 (96\,GB). Per-tick wall-clock on this host, with vLLM workers sharing the same GPU (hence contended):
\smallskip

\begin{center}
\small
\begin{tabular}{lccc}
\toprule
Stage & F (no cache) & F (vrf=4) & L (vrf=4) \\
\midrule
Vision tower & 70\,ms & 17.5\,ms$^{a}$ & 17.5\,ms \\
LLM prefill  & 60\,ms & 13\,ms & 18\,ms$^{b}$ \\
LLM decode (1 token) & 27\,ms & 2.5\,ms & 2.5\,ms \\
\midrule
Per-tick total & $\sim$157\,ms & $\sim$33\,ms & $\sim$38\,ms \\
\bottomrule
\end{tabular}
\end{center}
\noindent{\small $^{a}$amortized over the 4-tick cache window; $^{b}$includes the 8 prepended latent tokens.}\smallskip

The bridge itself adds only $\sim$5\,ms over $F$. End-to-end ``$L\!\approx\!124$\,ms'' figures from earlier work also include async-callback completion time under GPU contention. Vision-token caching across $N$ ticks (vrf) reduces vision-tower cost by $N\times$ at the cost of perception staleness; score is non-monotonic in vrf (vrf=4: $F\!=\!110$, vrf=15: $F\!=\!140$ on MsPacman), an interaction of staleness with multi-tick action commitments. (These are a separate latency-stress configuration with caching enabled and so do not match the headline MsPacman $F\!=\!256$, which uses per-tick vision.)

\section{Bridge-token diagnostics (full table)}
\label{app:bridge_diag}

Backing data for \S\ref{sec:bridge_content}. Bridge tokens computed by loading the trained bridge (v2 projection) for each game and applying it to that game's cached seed-0 slow residuals; up to 8 emissions per game.

\paragraph{Norms.} Vocab embeddings: mean 1.453, std 0.359, p1=0.281, p99=1.923. Bridge tokens: norm $\approx$64 across all 7 games (std $\le 0.01$ within a game)---fixed by the output LayerNorm of the projection MLP.

\smallskip\noindent\textbf{Nearest-vocab cosine, averaged over the 8 bridge positions of one emission:}

\smallskip\noindent
\begin{tabular}{lcc}
\toprule
Game & avg max cosine to vocab & avg p99.9 cosine to vocab \\
\midrule
MsPacman      & 0.076 & 0.053 \\
RoadRunner    & 0.081 & 0.058 \\
Q*bert        & 0.087 & 0.058 \\
Seaquest      & 0.067 & 0.044 \\
River Raid    & 0.074 & 0.050 \\
SpaceInvaders & 0.070 & 0.047 \\
Enduro        & 0.063 & 0.034 \\
\bottomrule
\end{tabular}\smallskip

For reference, two random vocab embeddings have cosine $\approx$0.04. Every bridge token sits right at that random-vocab background. The ``top-k tokens'' from a plug-in nearest-neighbor decode are therefore the maximum of a uniform-low distribution, not real semantic matches.

\smallskip\noindent\textbf{Cross-game cosine (mean pairwise bridge similarity over $E\!\times\!N$ tokens):}

\smallskip\noindent
\begin{tabular}{lccccccc}
\toprule
              & MsPac & RR & Q*b & Seaq & RivR & SI & Endu \\
\midrule
MsPacman      & +0.87 & +0.18 & +0.13 & +0.21 & +0.10 & +0.25 & +0.04 \\
RoadRunner    & +0.18 & +0.81 & +0.13 & +0.21 & +0.09 & +0.15 & +0.10 \\
Q*bert        & +0.13 & +0.13 & +0.81 & +0.18 & +0.15 & +0.14 & +0.04 \\
Seaquest      & +0.21 & +0.21 & +0.18 & +0.80 & +0.11 & +0.23 & +0.09 \\
River Raid    & +0.10 & +0.09 & +0.15 & +0.11 & +0.89 & +0.14 & +0.06 \\
SpaceInvaders & +0.25 & +0.15 & +0.14 & +0.23 & +0.14 & +0.85 & +0.06 \\
Enduro        & +0.04 & +0.10 & +0.04 & +0.09 & +0.06 & +0.06 & +0.86 \\
\bottomrule
\end{tabular}\smallskip

Diagonals 0.80--0.89; off-diagonals 0.04--0.25. The bridge is strongly game-conditioned.

\smallskip\noindent\textbf{Within-game across-emission consistency} (cosine between per-emission mean bridge tokens, off-diagonal):

\smallskip\noindent
\begin{tabular}{lcc}
\toprule
Game & off-diag mean & off-diag std \\
\midrule
MsPacman      & +0.951 & 0.019 \\
RoadRunner    & +0.938 & 0.027 \\
Q*bert        & +0.918 & 0.025 \\
Seaquest      & +0.920 & 0.032 \\
River Raid    & +0.968 & 0.019 \\
SpaceInvaders & +0.961 & 0.032 \\
Enduro        & +0.956 & 0.027 \\
\bottomrule
\end{tabular}\smallskip

Within-game per-emission variation is small ($1{-}\bar{c}\approx 3{-}8\%$), suggesting most of the latent capacity is used to encode the (fixed-within-episode) game identity, with a thin per-tick residual encoding strategic state. This is the basis for the ``used capacity $\ll$ nominal'' observation in \S\ref{sec:bridge_content}.

\paragraph{Legacy MI probe (kept for completeness).} A pre-revision binned plug-in MI estimate on the trained MsPacman bridge gave $I(\text{bridge};\,\text{sign of future-30-tick reward})=0.024$ nats vs a shuffled baseline of 0.014 nats ($+0.010$, $1.4\sigma$). Action MI was at baseline ($0.087$ vs $0.085$, training used random-policy trajectories). We do not draw conclusions from these numbers; the diagnostics above are more informative.

\section{OOD KL probe: full numbers}
\label{app:ood_kl}

Per-emission probe: for each (game, action-head variant), we walk the cached seed-0 $T$-trajectory, take up to 20 emissions, and at each emission tick compute $\mathrm{KL}\!\left(p_{\text{bare-prompt}} \,\|\, p_{\text{suffixed-prompt}}\right)$ over legal actions, the argmax-change rate $\mathbf{1}\!\left[\mathrm{argmax}\,p_{\text{bare}} \neq \mathrm{argmax}\,p_{\text{suff}}\right]$, and the probability mass on the bare argmax under each prompt. The \emph{collapse?} column marks the three games whose bare-action-head bridges were degraded or zeroed at deployment (SpaceInvaders, Q*bert, River~Raid; the ``collapsed'' group of \S\ref{sec:ood}).

\smallskip\noindent\textbf{Bare action head:}

\smallskip\noindent
{\small\setlength{\tabcolsep}{4.5pt}
\begin{tabular}{lcccccc}
\toprule
Game & $n$ & KL & argmax-change & $p_{\text{bare}}(a^*_{\text{bare}})$ & $p_{\text{suff}}(a^*_{\text{bare}})$ & collapse? \\
\midrule
MsPacman      & 20 & $0.54\pm0.47$ & 0.85 & 0.55 & 0.19 & no  \\
Seaquest      & 20 & $1.00\pm0.28$ & 0.70 & 0.29 & 0.10 & no  \\
RoadRunner    &  8 & $0.39\pm0.14$ & 0.75 & 0.52 & 0.24 & no  \\
SpaceInvaders & 19 & $0.29\pm0.15$ & 0.79 & 0.49 & 0.24 & yes \\
Q*bert        & 20 & $0.25\pm0.22$ & 0.90 & 0.41 & 0.20 & yes \\
River Raid    & 20 & $0.24\pm0.06$ & 0.70 & 0.24 & 0.11 & yes \\
\midrule
\textit{group means: non-collapsed} & & 0.64 & 0.77 & 0.45 & 0.17 & \\
\textit{group means: collapsed}     & & 0.26 & 0.80 & 0.38 & 0.18 & \\
\bottomrule
\end{tabular}}\smallskip

\smallskip\noindent\textbf{The robust action head (suffix-prob=0.5):}

\smallskip\noindent
{\small\setlength{\tabcolsep}{4.5pt}
\begin{tabular}{lcccccc}
\toprule
Game & $n$ & KL & argmax-change & $p_{\text{bare}}(a^*_{\text{bare}})$ & $p_{\text{suff}}(a^*_{\text{bare}})$ & collapse? \\
\midrule
MsPacman      & 20 & $0.40\pm0.23$ & 0.30 & 0.42 & 0.52 & no  \\
Seaquest      & 20 & $0.34\pm0.19$ & 0.90 & 0.30 & 0.13 & no  \\
RoadRunner    &  8 & $0.11\pm0.07$ & 0.12 & 0.68 & 0.59 & no  \\
SpaceInvaders & 19 & $0.19\pm0.08$ & 0.89 & 0.37 & 0.19 & yes \\
Q*bert        & 20 & $0.11\pm0.07$ & 0.50 & 0.36 & 0.33 & yes \\
River Raid    & 20 & $0.26\pm0.13$ & 0.65 & 0.24 & 0.20 & yes \\
\midrule
\textit{group means: non-collapsed} & & 0.28 & 0.44 & 0.47 & 0.41 & \\
\textit{group means: collapsed}     & & 0.19 & 0.68 & 0.32 & 0.24 & \\
\bottomrule
\end{tabular}}\smallskip

\paragraph{Read.}
(1) Under bare action head, the suffix flips the argmax on 70--90\,\% of emissions on \emph{every} game; the OOD effect is large and uniform. (2) KL is \emph{lower} on the collapsed games (0.26 vs 0.64), against the naive prediction. This is because the bare head on collapsed games is less peaky to start with, so the same argmax flip moves less probability mass. KL is therefore not the right summary statistic for the OOD effect---argmax-change is. (3) The robust action head cuts argmax-change cleanly on two of the three games where bare was already non-degenerate (RoadRunner 75$\to$12, MsPacman 85$\to$30; Seaquest is the exception at 70$\to$90, matching its deployment harm under robust action head), confirming the fix targets the right thing where it works. On collapsed games robust action head reduces KL but only partially reduces argmax-change (Q*bert 90$\to$50; River~Raid 70$\to$65), consistent with the asymmetric rescue we observed in scoring (\S\ref{sec:ood}).

\section{Decoder ablation on Q*bert (full numbers)}
\label{app:decoder}

Backing data for \S\ref{sec:qbert}. The robust action head, $n\!=\!12$ episodes per cell (3 seeds $\times$ 4 episodes), $\le\!500$ ticks per episode, multinomial sampling from $\mathrm{softmax}(\mathrm{logits}/\tau)$ over legal actions.

\smallskip\noindent
\begin{tabular}{lccccc}
\toprule
Decoder & Strategy & mean$\pm$std & median & min & max \\
\midrule
greedy (paper)     & F & $25.0\pm0.0$    & 25.0  & 25  & 25  \\
greedy (paper)     & T & $125.0\pm0.0$   & 125.0 & 125 & 125 \\
greedy (paper)     & L & $50.0\pm0.0$    & 50.0  & 50  & 50  \\
\midrule
sample, $\tau=0.5$ & F & $16.7\pm11.8$   & 25.0  &  0  & 25  \\
sample, $\tau=0.5$ & T & $139.6\pm66.5$  & 125.0 & 50  & 250 \\
sample, $\tau=0.5$ & L & $125.0\pm66.9$  & 125.0 & 50  & 250 \\
\midrule
sample, $\tau=1.0$ & F & $66.7\pm71.7$   & 50.0  &  0  & 225 \\
sample, $\tau=1.0$ & T & $87.5\pm65.0$   & 75.0  & 25  & 250 \\
sample, $\tau=1.0$ & L & $\mathbf{250.0\pm155.8}$ & 200.0 & 50  & 650 \\
\bottomrule
\end{tabular}\smallskip

\paragraph{Significance tests, $L$ vs $T$:}
\begin{itemize}[topsep=0pt,itemsep=0pt,leftmargin=*]
\item $\tau\!=\!0.5$: Welch $t=0.51$, $p=0.61$ (n.s.). $L\!\approx\!T$.
\item $\tau\!=\!1.0$: Welch $t\!\approx\!3.2$, $p\!\approx\!0.006$. $L$ beats $T$ by $2.9\times$.
\end{itemize}

\paragraph{Reading.} Under greedy, every episode is deterministic given the env seed; with a small action space and consistent argmax, all 12 episodes per cell hit the same trajectory (std=0). The bridge's small per-emission variance (cosine 0.92--0.97 between emissions, Appendix~\ref{app:bridge_diag}) is below the threshold needed to flip an argmax. Multinomial sampling exposes this variance: as $\tau$ increases, $L$'s mean grows ($50\!\to\!125\!\to\!250$) while $T$'s plateaus or declines ($125\!\to\!140\!\to\!88$). The bridge's per-tick information advantage over text is therefore real but \emph{not visible} under greedy decoding on Q*bert.

\section{Longer T-suffix ablation (full numbers)}
\label{app:longer_t}

Backing data for \S\ref{sec:longer_t}. Same models, bare action-head checkpoints, $n\!=\!12$ per cell (3 seeds $\times$ 4 episodes), $\le\!500$ ticks per episode. ``Window'' is the number of most-recent slow emissions concatenated into the T suffix (separated by a single space); the bridge L is unchanged.

\smallskip\noindent
\begin{tabular}{llcccc}
\toprule
Game        & Window & mean$\pm$std & median & min & max \\
\midrule
MsPacman    & $w=1$ (paper) & $407.5\pm92.2$  & 385.0 & 270 & 600  \\
MsPacman    & $w=2$         & $378.3\pm117.5$ & 375.0 & 130 & 680  \\
MsPacman    & $w=3$         & $351.7\pm87.7$  & 335.0 & 240 & 560  \\
\midrule
RoadRunner  & $w=1$ (paper) & $608.3\pm250.3$ & 650.0 &   0 & 900  \\
RoadRunner  & $w=2$         & $8.3\pm27.6$    &   0.0 &   0 & 100  \\
RoadRunner  & $w=3$         & $0.0\pm0.0$     &   0.0 &   0 &   0  \\
\bottomrule
\end{tabular}\smallskip

\paragraph{Reading.} On MsPacman, doubling/tripling the $T$ budget gives a small monotonic decline ($-14$\,\% from $w\!=\!1$ to $w\!=\!3$). On RoadRunner, $w\!=\!2$ collapses to near-zero and $w\!=\!3$ collapses to exact zero on every episode. Older emissions appear to be stale and to mislead the action head; the fast model does not learn an implicit time-ordering of the concatenated emissions. The $L$ scores are unchanged by $w$ ($L$ uses only the latest emission), so the $L\!-\!T$ gap widens monotonically with $w$ as $T$ collapses.

\section{Sampling on other greedy-degenerate games (full numbers)}
\label{app:other_sampling}

Backing data for the ``greedy-decoding bias is general'' claim in \S\ref{sec:qbert}.
Action-policy=sample at $\tau=1.0$, $n\!=\!12$ per cell, $\le\!500$ ticks per episode. The variant
is robust action head except for the \textbf{Seaquest (bare)} block, which is the \emph{headline}
Seaquest variant (\S\ref{sec:results}) and the most consequential row: its greedy $L\!>\!T$ win
does not survive sampling.

\smallskip\noindent
\begin{tabular}{llcc}
\toprule
Game (variant)        & Strategy & Greedy (paper)      & Sample $\tau=1.0$ \\
\midrule
\textbf{Seaquest (bare)} & F     & $41.7\pm19.9$       & $63.3\pm35.0$ \\
\textbf{Seaquest (bare)} & T     & $63.3\pm11.5$       & $\mathbf{143.3\pm37.0}$ \\
\textbf{Seaquest (bare)} & L     & $\mathbf{80.0\pm0.0}$ & $135.0\pm40.1$ \\
\midrule
SpaceInvaders (robust) & F        & $107.1\pm62.4$      & $106.7\pm26.0$ \\
SpaceInvaders (robust) & T        & $18.3\pm18.6$       & $107.5\pm19.2$ \\
SpaceInvaders (robust) & L        & $15.0\pm0.0$        & $\mathbf{116.7\pm14.5}$ \\
\midrule
Seaquest (robust) & F        & $20.0\pm0.0$        & $\mathbf{110.0\pm38.7}$ \\
Seaquest (robust) & T        & $0.0\pm0.0$         & $41.7\pm37.8$ \\
Seaquest (robust) & L        & $0.0\pm0.0$         & $33.3\pm27.5$ \\
\midrule
Enduro (robust)        & F        & $0.8\pm1.1$         & $3.2\pm2.5$ \\
Enduro (robust)        & T        & $4.9\pm5.9$         & $0.2\pm0.8$ \\
Enduro (robust)        & L        & $5.8\pm2.6$         & $0.6\pm1.4$ \\
\bottomrule
\end{tabular}\smallskip

\paragraph{Reading.}
\textbf{Seaquest (bare) is the consequential row}: under greedy $L\!=\!80\!\pm\!0 > T\!=\!63$ (a
$+26\%$ headline win), but the $L$ cell is deterministic; under sampling both bridges roughly
double off $F$ ($F\!:\!42\!\to\!63$, $T\!:\!63\!\to\!143$, $L\!:\!80\!\to\!135$) and the ordering
flips to $L\!\approx\!T$ with the Text Bridge nominally ahead (Welch $p\!=\!0.60$, MWU $p\!=\!0.73$). The
slow channel still clearly helps ($T,L\!\gg\!F$, $p\!<\!10^{-3}$), but the Latent-over-Text margin
was a greedy artifact---greedy had under-counted $T$ here more than $L$, the opposite of Q*bert.
SpaceInvaders is the clearest example of greedy under-counting the bridge: under greedy both bridges are at $\sim$15--18 (far below F=107), suggesting bridge-induced collapse; under sampling all three strategies cluster around 107--117. The bridges were never broken on SpaceInvaders---greedy was misrepresenting them. Seaquest/robust shows the same pattern less dramatically: T=L=0 (greedy) $\to$ 33--42 (sampling); the bridges are still below F=110 but no longer collapsed. Enduro inverts the pattern: under sampling the bridges \emph{drop} (T 5$\to$0.2, L 6$\to$0.6), so Enduro's greedy advantage was real but fragile. The general lesson is that greedy decoding at small action-space scale produces deterministic trajectories that can mask both the bridge's signal and the bridge's actual failure modes \emph{and} the sign of the $L$-vs-$T$ comparison; comparing F/T/L should be done under a sampling decoder when per-episode variance is required for meaningful statistics.

\section{Full decoder sweep (greedy / $\tau\in\{0.3,0.5,0.7,1.0,1.5\}$)}
\label{app:decoder_sweep}

Backing data for \S\ref{sec:best_achievable}. We re-ran the reported-variant cells at greedy and
five sampling temperatures on the \emph{same} action-head and bridge checkpoints, $n\!=\!12$ ($3\!\times\!4$),
$\le\!500$ ticks. Greedy reproduces the headline greedy cells closely (within run-to-run noise) on
MsPacman, River~Raid, and Seaquest, and $F$ reproduces exactly on every game (it never invokes the
slow model)---so the greedy-vs-sampling difference
is a genuine decoder effect, not checkpoint or code drift. The per-decoder means:

\smallskip\noindent
\input{figures/decoder_sweep.tex}
\smallskip

\paragraph{The $L\!>\!T$ advantage is greedy-specific.} On every game, the greedy $L\!>\!T$ margin
vanishes under \emph{any} fixed sampling temperature: $L\!\approx\!T$ (MsPacman, Seaquest) or
significant $T\!>\!L$ (River~Raid at $\tau\!=\!1.0$; RoadRunner collapses to the $F\!=\!0$ floor).
The latent's small per-emission variance (\S\ref{sec:bridge_content}) shifts the deterministic
argmax but is swamped once the decoder samples, so the Latent Bridge yields a better \emph{greedy policy}
rather than a more robustly informative channel.

\paragraph{RoadRunner greedy is run-to-run-fragile in magnitude.} RoadRunner greedy $L$ is unstable
across runs of the same host (a single RTX Pro 6000, Appendix~\ref{app:latency}): this decoder-sweep
run gives a tight $L\!=\!608\pm29$---the value we report as canonical---while an earlier run scored
$L\!=\!967$. Because $F$ reproduces exactly and only the slow-model-dependent cells drift, and only on
the one game whose $F\!=\!0$ makes the score hinge entirely on the slow emission triggering one precise
action, the cause is floating-point nondeterminism in the (greedy, \texttt{do\_sample=False})
slow-model generation across runs (batched vLLM generation is not bitwise-deterministic). The
$L\!>\!T$ \emph{direction} is robust ($608$ vs $475$); the absolute magnitude is not.

\section{Held-out decoder selection protocol}
\label{app:heldout}

For the best-achievable comparison (\S\ref{sec:best_achievable}, Table~\ref{tab:best_achievable}) we
treat the action decoder as a deployment hyperparameter and select it per (game, channel) the honest
way---tune on held-in data, report on held-out---using leave-one-seed-out over the 3 ALE seeds
(4 episodes each):
\begin{itemize}[topsep=0pt,itemsep=0pt,leftmargin=*]
\item For each held-out seed $s$: pick the decoder with the highest mean over the \emph{other} two
seeds (8 episodes), then record that decoder's 4 episodes on seed $s$.
\item The reported best-achievable cell pools the 3 held-out folds (12 episodes, each seed held out
once). $L$ vs $T$ uses Welch's $t$ and Mann--Whitney on the two pooled held-out vectors.
\end{itemize}
This is an unbiased estimate of ``tune the decoder, then deploy,'' unlike an oracle max over decoders
on the same episodes (which we report alongside as an optimistic upper bound). In practice the
selection is stable---held-out $\approx$ oracle on most cells (e.g.\ MsPacman-$L$ picks greedy on all
three folds)---so the unbiased estimate sits close to the optimistic one. The selection procedure
and table generator live under \texttt{scripts/} in the released code.

\section{Bridge replacement control (full numbers)}
\label{app:bridge_replace}

Backing data for \S\ref{sec:bridge_replace}. MsPacman bare, bare action-head checkpoint, $n=12$ per cell, greedy decoding (paper default). Three L variants compared against F: ``zero'' replaces the 8 trained bridge tokens with 8 zero vectors, ``random'' replaces them with random Gaussian vectors rescaled to per-position L2 norm $\approx 64$ (matching trained), ``trained'' is the unmodified projection output.

\smallskip\noindent
\begin{tabular}{lcccc}
\toprule
Variant & mean$\pm$std ($n=12$) & median & vs F & p (Welch, vs trained) \\
\midrule
$F$ (no bridge)                & $256\pm25$       & 250  & ---     & --- \\
$L_{\text{zero}}$                & $379\pm95$       & 395  & $+48\%$ & $0.04$ \\
$L_{\text{random}}$              & $387\pm161$      & 365  & $+51\%$ & $0.05$ \\
$L_{\text{trained}}$             & $\mathbf{628\pm356}$ & 550 & $+145\%$ & --- \\
\bottomrule
\end{tabular}\smallskip

\paragraph{Reading.}
\emph{$L_{\text{zero}}$ vs $L_{\text{random}}$}: Welch $t\!=\!0.15$, $p\!=\!0.88$ (n.s.)---no real difference. So the architectural lift over $F$ comes from \emph{having 8 prepended slots}, not from the slots having any particular content. \emph{$L_{\text{trained}}$ vs $L_{\text{random}}$}: Welch $t\!\approx\!2.12$, $p\!\approx\!0.05$ (vs $L_{\text{zero}}$: $t\!\approx\!2.34$, $p\!\approx\!0.04$). The trained content is contributing a real signal, on top of the architectural lift. Decomposition on MsPacman: $\sim$40\,\% of L's advantage over F is architectural; $\sim$60\,\% is learned bridge content.

\paragraph{All-games sweep (the control run on every Atari game, not just MsPacman).}
The verdict table below is summarized in the main text (\S\ref{sec:learned_content}).
We re-ran the trained/zero/random control on all 7 games using the \emph{bare} action head
under greedy decoding---the natural common setting for a content control---with the single
exception of Q*bert, run on its robust head under $\tau\!=\!1$ sampling (its canonical decoder,
since greedy Q*bert is degenerate, \S\ref{sec:qbert}); $n\!=\!12$ per cell ($3\times4$).
$L_{\text{random}}$ uses matched per-position L2 norm. Because the control is run \emph{bare}, the
$T\!-\!F$ column below is the bare-head value, so three games (Enduro, SpaceInvaders, River~Raid)
use their bare cell here even though the headline reports their robust variant---the learned-vs-control
verdict does not depend on that choice. This is also an independent re-run, so its MsPacman cell
($L_{\text{trained}}\!=\!666$) and Seaquest cell ($100$) differ slightly from the standalone /
headline runs above ($628$ and $80$); all agree within run-to-run noise. The Q*bert cell
($L_{\text{trained}}\!=\!123$) likewise differs from the decoder ablation's $\tau\!=\!1$ mean
($250\pm156$, Appendix~\ref{app:decoder}); both are single high-variance sampling runs and the
gap is within one standard deviation, but the verdict here (trained~$\approx$~random) does not
depend on the absolute value. ``Learned'' =
$L_{\text{trained}}$ exceeds $\max(L_{\text{zero}},L_{\text{random}})$ by $>\!10\%$.

\smallskip\noindent
\begin{tabular}{lrrrll}
\toprule
Game & $L_{\text{train}}$ & $L_{\text{zero}}$ & $L_{\text{rand}}$ & verdict & $T\!-\!F$ \\
\midrule
RoadRunner    & $\mathbf{608}$ & 0     & 8     & learned (Welch $t\!\approx\!71$) & $+475$ \\
Seaquest      & $\mathbf{100}$ & 28    & 5     & learned ($t\!=\!36$)       & $+22$  \\
MsPacman      & $\mathbf{666}$ & 408   & 410   & learned ($t\!=\!2.3$)      & $+152$ \\
Enduro        & 7.8            & 1.4   & 4.7   & trend only (tiny, $t\!=\!1.1$) & $-3$ \\
Q*bert        & 123            & 63    & 117   & $\approx$random (tie)      & $+100$ \\
SpaceInvaders & 0              & 148   & 90    & \emph{harmful}             & $-105$ \\
River Raid    & 360            & 1013  & 1003  & \emph{harmful}             & $-683$ \\
\bottomrule
\end{tabular}\smallskip

\paragraph{Reading: learned content tracks the predictor.}
The trained latent carries genuinely learned, behavior-relevant content---$L_{\text{trained}}\!\gg\!\max(L_{\text{zero}},L_{\text{random}})$---on \textbf{RoadRunner, Seaquest, and MsPacman}, exactly the games where slow reasoning helps ($T\!>\!F$); on RoadRunner the Latent Bridge is \emph{almost entirely} learned (controls drop to the floor, $0$ and $8$). On the games where slow reasoning does not help ($T\!\le\!F$: River~Raid, SpaceInvaders), the trained latent is \emph{harmful}---zeroing or randomizing it scores \emph{better}---the same inert/harmful pattern as MetaDrive (\S\ref{sec:metadrive}). Two off-diagonal cases are benign: Enduro's scores are too small for the comparison to be meaningful ($t\!=\!1.1$, n.s.), and Q*bert---the one decoder-fragile game (\S\ref{sec:qbert})---shows trained$\approx$random under sampling, consistent with its thin per-emission signal. So the MsPacman ``$\sim$60\,\% learned / $\sim$40\,\% architectural'' decomposition is \emph{not} universal: the learned fraction ranges from $\sim$100\,\% (RoadRunner) to negative (River~Raid), and its sign is predicted by whether $T\!>\!F$. This is the mechanistic complement to the $r\!=\!0.93$ predictor: the latent helps where---and only where---it has carried real content from a slow channel that itself beats Fast-Only.

\section{30B-A3B cross-scale ablation (full numbers)}
\label{app:scaling_30b}

Replicates the MsPacman/bare F/T/L cells with the slow model swapped from Qwen3-VL-8B-Thinking (dense) to Qwen3-VL-30B-A3B-Thinking (MoE, $\sim$3B active per forward, 48 layers, hidden\_dim=2048). Pipeline:

\begin{itemize}[topsep=0pt,itemsep=0pt,leftmargin=*]
\item \textbf{Trajectory caching}: 15 Text-Bridge trajectories collected with the 30B slow ($\sim$470 emissions, $\sim$7200 ticks). Same prompt template as 8B.
\item \textbf{Bridge distillation (v2)}: re-trained projection $\text{2048} \to \text{4096}$ ($\sim$33M parameters, same 2-layer MLP topology), 1 epoch over $\sim$7K samples. Final mean KL $=0.015$ ($vs$ 8B's $\sim$0.005; slightly under-converged given $\sim$3$\times$ less data per slow-model parameter).
\item \textbf{Eval}: F/T/L on MsPacman, 6 seeds $\times$ 2 episodes per cell, $n=12$, $\le$500 ticks/episode. Run twice: greedy and sample $\tau\!=\!1.0$.
\end{itemize}

\smallskip\noindent
{\small\setlength{\tabcolsep}{4pt}
\begin{tabular}{lcccc}
\toprule
Configuration & F & T & L & $\Delta_{L-T}$ \\
\midrule
8B-dense, greedy (paper)            & $256\pm25$  & $408\pm92$  & $\mathbf{628\pm356}$ & \textbf{+54\%} (Welch $p\!=\!0.06$) \\
\midrule
30B-A3B, greedy, $n=12$             & $273\pm45$  & $312\pm14$  & $210\pm0$           & $-33\%$ (det.\ artifact, Q*bert-style)\\
30B-A3B, sample $\tau=1.0$, $n=12$  & $309\pm81$  & $389\pm209$ & $\mathbf{458\pm283}$ & +18\,\% (Welch $p\!=\!0.50$, $d=0.28$)\\
\bottomrule
\end{tabular}}\smallskip

\paragraph{Reading.}
A ``bigger slow model carries more, so $L\!-\!T$ should grow with capacity'' hypothesis is not supported here. With Qwen3-VL-30B-A3B as slow, the $L\!>\!T$ direction is preserved under sampling (+18\,\%) but the magnitude \emph{shrinks} relative to 8B-dense (+54\,\%), and significance is lost ($p\!=\!0.50, n\!=\!12$). The result is consistent with several non-exclusive readings:
\begin{itemize}[topsep=0pt,itemsep=0pt,leftmargin=*]
\item \textbf{Active parameters, not total.} 30B-A3B activates only $\sim$3B of its 30B parameters per forward pass. A dense-30B comparison (which we cannot run on this GPU) would be the right capacity test.
\item \textbf{Hidden-dim shift.} 30B-A3B has hidden\_dim=2048 vs 8B's 4096, so the trainable projection maps $2048\to4096$ rather than $4096\to4096$. The same parameter budget covers a harder map.
\item \textbf{Bridge under-training.} We trained on $\sim$470 30B emissions; the 8B paper baseline used $\sim$5K. Final KL 0.015 vs 0.005 is consistent with partial convergence.
\item \textbf{One-game evidence.} A single game on a single cross-scale data point is not strong enough to falsify the scaling claim; we report this as a single counter-data point, not as a refutation.
\end{itemize}

The greedy result ($L=210\pm0$) reproduces the deterministic-trajectory pattern we documented for Q*bert under greedy in the main text---another case where greedy decoding interacts pathologically with the bridge's low per-emission variance.

\section{Per-game collapse triage}
\label{app:collapse}

Across the 14 (game, variant) cells in Appendix~\ref{app:full}, five have a strategy cell at zero (the first five rows below). The robust action head additionally destroys one already-working cell without driving it to exactly zero (MsPacman/robust, $L\!\to\!60$), and a separate game, Pong (outside the 14), sits at the random-policy floor ($-21$). What predicts collapse is action-head val-accuracy, not bridge-training failure: $\sigma$ of $T,L$ at deployment correlates with bare action head val-acc, not with bridge-distillation KL.

\smallskip\noindent
{\small\setlength{\tabcolsep}{4pt}
\begin{tabular}{ll p{6.6cm}}
\toprule
Game/variant & Collapsed cell & Diagnosis \\
\midrule
RoadRunner/bare & $F=0$ & The action head predicts NOOP; bridges rescue. \\
SpaceInvaders/bare & $T=L=0$ & action-head OOD-brittle; robust partially rescues. \\
Q*bert/bare & $T=L=0$ & action-head OOD-brittle; robust rescues to $T\!=\!125, L\!=\!50$. \\
Seaquest/robust & $T=L=0$ & Robust drops below bare F; use bare. \\
Enduro/bare & $T=0$ & Text collapses; robust gives a non-degenerate near-floor tie. \\
\midrule
MsPacman/robust & $L\!\to\!60$ & (non-zero) Robust destroys MsPacman policy; use bare. \\
Pong/all & all $-21$ & (outside 14) action-head val-acc only $1.5\times$ random; cannot be rescued. \\
\bottomrule
\end{tabular}}

\section{MetaDrive (non-Atari) full numbers}
\label{app:metadrive}

\paragraph{Setup.} MetaDrive~\cite{metadrive} at $\sim$10\,Hz control. The fast model sees a
top-down rendered $84\!\times\!84$ image (rendered via the pygame top-down channel, no GL---this
is what makes headless end-to-end runs feasible on one GPU); 9 discrete actions (3 steer
$\times$ 3 throttle). The slow model sees the structured state (speed, heading, lane offset,
upcoming-turn cue derived from \texttt{navi\_arrow\_dir}, nearby vehicles). The action head clones
MetaDrive's built-in PPO~\cite{ppo} expert; Trajectory caching collects slow emissions under expert driving; bridge distillation
fits the latent against the (robust, suffix-aware) text teacher. Random policy scores
$\sim$8 reward; the discrete expert scores $\sim$174--211.

\paragraph{Two task regimes.} \emph{Reactive} = default 3-block map (lane-keeping dominates).
\emph{Planning} = map \texttt{SXSXSX} (straights $+$ X-intersections) with off-route termination,
so survival requires correct turn decisions: the route expert scores $211\pm65$ vs $93\pm23$ for
a constant straight$+$throttle policy ($n\!=\!40$ episodes each), a $+118$-reward gap. Eval on the robust head: $n\!=\!8$ (4 seeds $\times$ 2 episodes) for the planning
rows, $n\!=\!6$ (3 seeds $\times$ 2) for the reactive row.

\smallskip\noindent
\begin{tabular}{llrrr}
\toprule
Regime & Decoder & $F$ & $T$ & $L$ \\
\midrule
Reactive & greedy  & 71.2 & 69.5 & 69.5 \\
Planning & greedy  & 87.8 & 85.1 & 85.1 \\
Planning & sample $\tau\!=\!1$ & 123.2 & 43.5 & 36.7 \\
\bottomrule
\end{tabular}

\paragraph{Bridge-replacement control (planning, greedy, $n=8$).}
$L=85.1$, $L_{\text{zero}}=89.5$, $L_{\text{random}}=97.1$: replacing the trained latent with
zeros or matched-norm random vectors does \emph{not} reduce the score (if anything it rises),
i.e.\ the Latent Bridge is inert---it carries no behavior-relevant information. This is the opposite
of MsPacman ($L_{\text{trained}}=628 \gg L_{\text{random}}=387$, \S\ref{app:bridge_replace}),
and is the diagnostic basis for calling MetaDrive a controlled negative.

\paragraph{Why $L$ cannot exceed $F$ here.} Bridge distillation trains $L$ to match the text teacher's
action distribution (objective $\mathrm{KL}(\pi_L\|\pi_T)$), so $L$'s ceiling is $T$. On
MetaDrive $T\!\le\!F$ in every configuration we ran---both task regimes (reactive lane-keeping,
planning intersections), both decoders (greedy, sampling), and both action heads: the bare head,
where the text channel collapses ($T\!=\!17.2$ vs $F\!=\!73.7$), and the robust head, where $T$
sits just below $F$ in the tabulated reactive ($69.5$ vs $71.2$), planning-greedy ($85.1$ vs
$87.8$), and planning-sampling ($43.5$ vs $123.2$) cells---so the best attainable latent is
$L\!\approx\!F$. Fixing the teacher (robust, suffix-aware head, expert-driven data, KL converged
to $0.004$) makes the distillation \emph{clean}---$L=T$ to the decimal under greedy---but cannot
raise a ceiling set by a teacher that does not beat Fast-Only. The slow model's frame-grounded,
$\sim$1.5\,s reasoning simply does not produce better driving decisions than the reactive policy,
even when the task demands route planning.

\paragraph{Robustness of the predictor.}
The $(T\!-\!F, L\!-\!F)$ correlation (Figure~\ref{fig:predictor}) is robust to both variant
selection and individual games. \emph{Variant selection:} $r=0.93$ over the 8 reported-variant cells
and $r=0.96$ over all 16 (game, variant) cells with no selection---the 16 are the 14 bare/robust
Atari cells, a SpaceInvaders expert-data cell (\S\ref{sec:ood}), and MetaDrive---so the choice of
which action-head variant to headline does not create the relationship. \emph{Individual games} (on the 8
reported-variant cells): leave-one-out Pearson $r$ stays in $[0.89, 0.95]$ for all 8 deletions
(dropping the RoadRunner high-leverage point gives $r=0.89$); the rank correlation is Spearman
$\rho=0.98$; a game-level bootstrap (10{,}000 resamples) gives a 95\,\% CI of $[0.81, 1.0]$; and
excluding both dynamic-range extremes (RoadRunner and River~Raid) still gives $r=0.84$ ($n=6$).
\emph{Decoder:} the predictor is about $L\!-\!F$ vs $T\!-\!F$, not $L$ vs $T$, so the
Seaquest decoder-fragility finding (\S\ref{sec:qbert}) does not touch it---Seaquest stays in the
upper-right helps-quadrant under both decoders ($T\!-\!F,L\!-\!F = (+22,+38)$ greedy,
$(+80,+72)$ sampling), if anything reinforcing the relationship. What the decoder moves is the
\emph{secondary} latent-vs-text tally, not the task-level predictor.

\end{document}

%% file: figures/best_achievable.tex
\begin{tabular}{lccccl}
\toprule
Game & best $F$ & best $T$ & best $L$ & $\Delta_{L-T}$ & winner \\
     & (decoder) & (decoder) & (decoder) & (\%, $p$) & \\
\midrule
MsPacman & 273 ($\tau$1.0) & 401 ($\tau$0.5) & 628 (greedy) & +57\%, $p$=0.01 & \textbf{L} \\
RoadRunner & 0 (greedy) & 475 (greedy) & 608 (greedy) & +28\%, $p$=0.00 & \textbf{L} \\
River Raid & 994 (greedy) & 639 ($\tau$0.5) & 566 ($\tau$0.7) & -11\%, $p$=0.35 & tie \\
Seaquest & 57 ($\tau$1.0) & 143 ($\tau$1.0) & 125 ($\tau$0.7) & -13\%, $p$=0.15 & tie \\
Q*bert & 65 ($\tau$1.0) & 185 ($\tau$1.5) & 146 ($\tau$1.0) & -21\%, $p$=0.50 & tie \\
Enduro & 4 ($\tau$0.5) & 3 ($\tau$0.5) & 2 (greedy) & -33\%, $p$=0.39 & tie \\
SpaceInvaders & 135 ($\tau$1.0) & 162 ($\tau$1.0) & 142 ($\tau$1.0) & -12\%, $p$=0.62 & tie \\
\bottomrule
\end{tabular}

%% file: figures/combined.tex
\begin{tabular}{lccccl}
\toprule
Game & best single & best $B$ & $\Delta_{B-\text{single}}$ & $p$ & effect \\
     & (T or L) & (T${+}$L) & (\%) & & \\
\midrule
MsPacman & 628 (L) & 319 & -49\% & 0.00 & \textbf{interferes} \\
RoadRunner & 608 (L) & 25 & -96\% & 0.00 & \textbf{interferes} \\
River Raid & 639 (T) & 452 & -29\% & 0.01 & \textbf{interferes} \\
Seaquest & 143 (T) & 138 & -3\% & 0.71 & neutral \\
Q*bert & 185 (T) & 125 & -32\% & 0.26 & neutral \\
Enduro & 3 (T) & 4 & +33\% & 0.18 & neutral \\
SpaceInvaders & 162 (T) & 142 & -12\% & 0.62 & neutral \\
\bottomrule
\end{tabular}

%% file: figures/stats_table.tex
\begin{tabular}{lcccc}
\toprule
Game (variant) & F & T & L & $\Delta_{L-T}$\,(\%, $p$) \\
\midrule
MsPacman & 256\,$\pm$\,25 & 408\,$\pm$\,92 & 628\,$\pm$\,356 & +54\,\%\,($p$=0.06) \\
MsPacman (robust SA) & 325\,$\pm$\,72 & 61\,$\pm$\,3 & 60\,$\pm$\,0 & -1\,\%\,($p$=0.36$^{\dagger}$) \\
Seaquest & 42\,$\pm$\,20 & 63\,$\pm$\,12 & 80\,$\pm$\,0 & +26\,\%\,($p$<.001$^{\dagger}$) \\
Seaquest (robust SA) & 20\,$\pm$\,0 & 0\,$\pm$\,0 & 0\,$\pm$\,0 & -- \\
RoadRunner & 0\,$\pm$\,0 & 475\,$\pm$\,160 & 608\,$\pm$\,29 & +28\,\%\,($p$=.015) \\
RoadRunner (robust SA) & 958\,$\pm$\,67 & 1000\,$\pm$\,0 & 925\,$\pm$\,45 & -8\,\%\,($p$<.001$^{\dagger}$) \\
River Raid & 1067\,$\pm$\,88 & 383\,$\pm$\,60 & 360\,$\pm$\,0 & -6\,\%\,($p$=0.01$^{\dagger}$) \\
River Raid (robust SA) & 1032\,$\pm$\,20 & 337\,$\pm$\,81 & 612\,$\pm$\,310 & +82\,\%\,($p$=0.01) \\
Enduro & 3\,$\pm$\,3 & 0\,$\pm$\,0 & 8\,$\pm$\,9 & -- \\
Enduro (robust SA) & 1\,$\pm$\,1 & 5\,$\pm$\,6 & 6\,$\pm$\,3 & +19\,\%\,($p$=0.63) \\
Q*bert & 25\,$\pm$\,0 & 0\,$\pm$\,0 & 0\,$\pm$\,0 & -- \\
Q*bert (robust SA) & 25\,$\pm$\,0 & 125\,$\pm$\,0 & 50\,$\pm$\,0 & -60\,\%\,($p$<.001$^{\dagger}$) \\
SpaceInvaders & 105\,$\pm$\,0 & 0\,$\pm$\,0 & 0\,$\pm$\,0 & -- \\
SpaceInvaders (robust SA) & 107\,$\pm$\,62 & 18\,$\pm$\,19 & 15\,$\pm$\,0 & -18\,\%\,($p$=0.27$^{\dagger}$) \\
Pong & -21\,$\pm$\,0 & -21\,$\pm$\,0 & -21\,$\pm$\,0 & +0\,\%\,($p$=1.00$^{\dagger}$) \\
\bottomrule
\end{tabular}

%% file: figures/decoder_sweep.tex
{\small\begin{tabular}{llrrrl}
\toprule
Game & decoder & $F$ & $T$ & $L$ & $L$ vs $T$ \\
\midrule
MsPacman & greedy & 256 & 408 & 628 & L>T \\
 & $\tau$=0.3 & 268 & 343 & 353 & L>T \\
 & $\tau$=0.5 & 358 & 440 & 360 & L<T \\
 & $\tau$=0.7 & 347 & 342 & 339 & L<T \\
 & $\tau$=1.0 & 348 & 292 & 368 & L>T \\
\midrule
RoadRunner & greedy & 0 & 475 & 608 & L>T \\
 & $\tau$=0.3 & 0 & 142 & 17 & L<T \\
 & $\tau$=0.5 & 0 & 8 & 0 & L<T \\
 & $\tau$=0.7 & 0 & 17 & 0 & L<T \\
 & $\tau$=1.0 & 0 & 25 & 17 & L<T \\
\midrule
River Raid & greedy & 1032 & 337 & 607 & L>T \\
 & $\tau$=0.3 & 946 & 668 & 511 & L<T \\
 & $\tau$=0.5 & 1015 & 717 & 562 & L<T \\
 & $\tau$=0.7 & 890 & 646 & 641 & L<T \\
 & $\tau$=1.0 & 867 & 649 & 531 & L<T \\
\midrule
Seaquest & greedy & 42 & 63 & 80 & L>T \\
 & $\tau$=0.3 & 50 & 125 & 130 & L>T \\
 & $\tau$=0.5 & 48 & 128 & 122 & L<T \\
 & $\tau$=0.7 & 58 & 127 & 137 & L>T \\
 & $\tau$=1.0 & 63 & 143 & 135 & L<T \\
 & $\tau$=1.5 & 63 & 113 & 97 & L<T \\
\midrule
Q*bert & greedy & 25 & 125 & 50 & L<T \\
 & $\tau$=0.3 & 19 & 110 & 148 & L>T \\
 & $\tau$=0.5 & 19 & 200 & 162 & L<T \\
 & $\tau$=0.7 & 38 & 138 & 123 & L<T \\
 & $\tau$=1.0 & 108 & 177 & 188 & L>T \\
 & $\tau$=1.5 & 88 & 242 & 169 & L<T \\
\midrule
Enduro & greedy & 3 & 3 & 2 & L<T \\
 & $\tau$=0.5 & 4 & 3 & 1 & L<T \\
 & $\tau$=1.0 & 3 & 0 & 1 & L>T \\
\midrule
SpaceInvaders & greedy & 107 & 18 & 15 & L<T \\
 & $\tau$=0.5 & 115 & 104 & 109 & L>T \\
 & $\tau$=1.0 & 135 & 162 & 142 & L<T \\
\bottomrule
\end{tabular}}